\pdfoutput=1

\documentclass[11pt]{article}

\usepackage[preprint]{acl}

\usepackage{times}
\usepackage{latexsym}

\usepackage[T1]{fontenc}
\usepackage[T5]{fontenc}

\usepackage[utf8]{inputenc}

\usepackage{microtype}

\usepackage{inconsolata}

\usepackage{graphicx}
\usepackage{amsmath}

\usepackage{booktabs} 
\usepackage{tabularx} 
\usepackage{hyperref}       
\usepackage{url}            
\usepackage{xurl} 
\usepackage{amsfonts}       
\usepackage{nicefrac}       
\usepackage{microtype}      
\usepackage{xcolor}         
\usepackage{graphicx}       
\usepackage{subfig} 
\usepackage{lscape}
\usepackage{changepage} 
\usepackage{float}  %

\title{VN-MTEB: Vietnamese Massive Text Embedding Benchmark}

\author{Loc Pham$^{\spadesuit}$, 
Tung Luu$^{\spadesuit}$,
Thu Vo$^{\spadesuit}$,
Minh Nguyen$^{\clubsuit}$,
Viet Hoang$^{\spadesuit}$,\\
$^{\spadesuit}$ GreenNode AI, Singapore\\
$^{\clubsuit}$School of Electrical Engineering, International University, VNU-HCMC, Vietnam\\
\texttt{\{locpb, tunglq, thu, viethq5\}@greennode.ai, \{nntminh\}@hcmiu.edu.vn}
}

\begin{document}
\maketitle
\begin{abstract}

Vietnam ranks among the top countries in terms of both internet traffic and online toxicity. As a result, implementing embedding models for recommendation and content control duties in applications is crucial. However, a lack of large-scale test datasets, both in volume and task diversity, makes it tricky for scientists to effectively evaluate AI models before deploying them in real-world, large-scale projects. To solve this important problem, we introduce a Vietnamese benchmark, VN-MTEB for embedding models, which we created by translating a large number of English samples from the Massive Text Embedding Benchmark using our new automated framework. We leverage the strengths of large language models (LLMs) and cutting-edge embedding models to conduct translation and filtering processes to retain high-quality samples, guaranteeing a natural flow of language and semantic fidelity while preserving named entity recognition (NER) and code snippets. Our comprehensive benchmark consists of 41 datasets from six tasks specifically designed for Vietnamese text embeddings. In our analysis, we find that bigger and more complex models using Rotary Positional Embedding outperform those using Absolute Positional Embedding in embedding tasks. Datasets are available at HuggingFace: \href{https://huggingface.co/collections/GreenNode/vn-mteb-68871433f0f7573b8e1a6686}{VN-MTEB} 
\end{abstract}

\section{Introduction}

Recent advancements in Large Language Models (LLMs) \cite{grattafiori-2024-llama3herdmodels, deepseekai-2025-deepseekr1incentivizingreasoningcapability, gemma-team-2025-gemma3technicalreport} have led to significant improvements in various Natural Language Processing (NLP) tasks. To the best of our knowledge, numerous benchmarks have been established for NLP tasks; they predominantly focus on widely spoken languages such as English and Chinese \cite{muennighoff-etal-2023-mteb}. In contrast, low-resource languages like Vietnamese, which is spoken by over 100 million people \footnote{https://www.macrotrends.net/global-metrics/countries/vnm/vietnam/population}, have yet to benefit from the creation of large-scale benchmarks. Although several datasets have been published, including ViQuAD \cite{nguyen-etal-2020-vietnamese-vietquad}, ViMMRC \cite{van2020enhancing-ViMMRC}, and UIT-VSFC \cite{uit-vsfc}, these resources are often limited to a single task and domain, with a noticeable scarcity in their publication.

Text embedding methods \cite{cao-2024-recentadvancestextembedding} have become increasingly popular in both industrial and academic fields due to their critical role in a variety of natural language processing tasks. The significance of universal text embeddings has been further highlighted with the rise of LLMs applications such as Retrieval-Augmented Systems (RAGs) \cite{lewis-2021-retrievalaugmentedgenerationknowledgeintensivenlp}. Consequently, researchers who seek to evaluate models must often resort to manually collecting datasets and converting them into formats suitable for model evaluation, a process that is both time-consuming and labor-intensive. The Massive Text Embedding Benchmark (MTEB) \cite{muennighoff-etal-2023-mteb} was created to collect data and standardize ways to evaluate and score different text embedding models. However, for low-resource languages like Vietnamese, there is still a lack of diverse datasets covering various tasks and domains, as well as a standardized approach to benchmarking text embedding at scale.

Machine translation methods often require human intervention for quality verification \cite{qian-etal-2024-what-do-large-language-models-need-for-machine-translation-evaluation}, sample collection for benchmarks, and overall evaluation, leading to a significant increase in effort. To address this challenge, our approach integrates translation with additional quality assurance to ensure that our translated datasets satisfy key criteria. By utilizing the latest state-of-the-art models in text embedding, language detection, and LLMs for automatic translation and filtering of low-quality samples, we minimize the need for human intervention. This approach strikes a balance between high resource consumption (time, infrastructure) and high-quality output, with a significantly reduced human effort.

Recognizing the need for a standardized benchmark, this paper introduces VN-MTEB (Vietnamese Massive Text Embedding Benchmark). The scope and key contributions of this work are as follows.

\begin{itemize}

\item We introduce \textbf{VN-MTEB} - a substantial benchmark consisting of \textbf{41 datasets} from \textbf{6 tasks} (retrieval, reranking, classification, clustering, pair classification, and semantic textual similarity), designed to evaluate text embeddings for the Vietnamese language.
\item We contribute to and integrate with MTEB\footnote{https://huggingface.co/spaces/mteb/leaderboard} and make the source code used in the experiments available to the public.
\item We evaluate a collection of embedding models, including both multilingual and monolingual variants, on the VN-MTEB benchmark, and provide insights into the correlation between model types and their performance across various tasks.
\item We propose a translation method that enables strict control over the fidelity of synthesized samples by considering multiple evaluation criteria. The goal of this approach is to facilitate translation tasks without requiring human involvement in either the translation or the quality assurance process.
\end{itemize}


\section{Related Works}\label{sec:related_works}
\subsection{Benchmarks and MTEB}\label{subsec:benchmarks}
\begin{figure*}[ht!]
    \centering
    \includegraphics[width=0.8\textwidth]{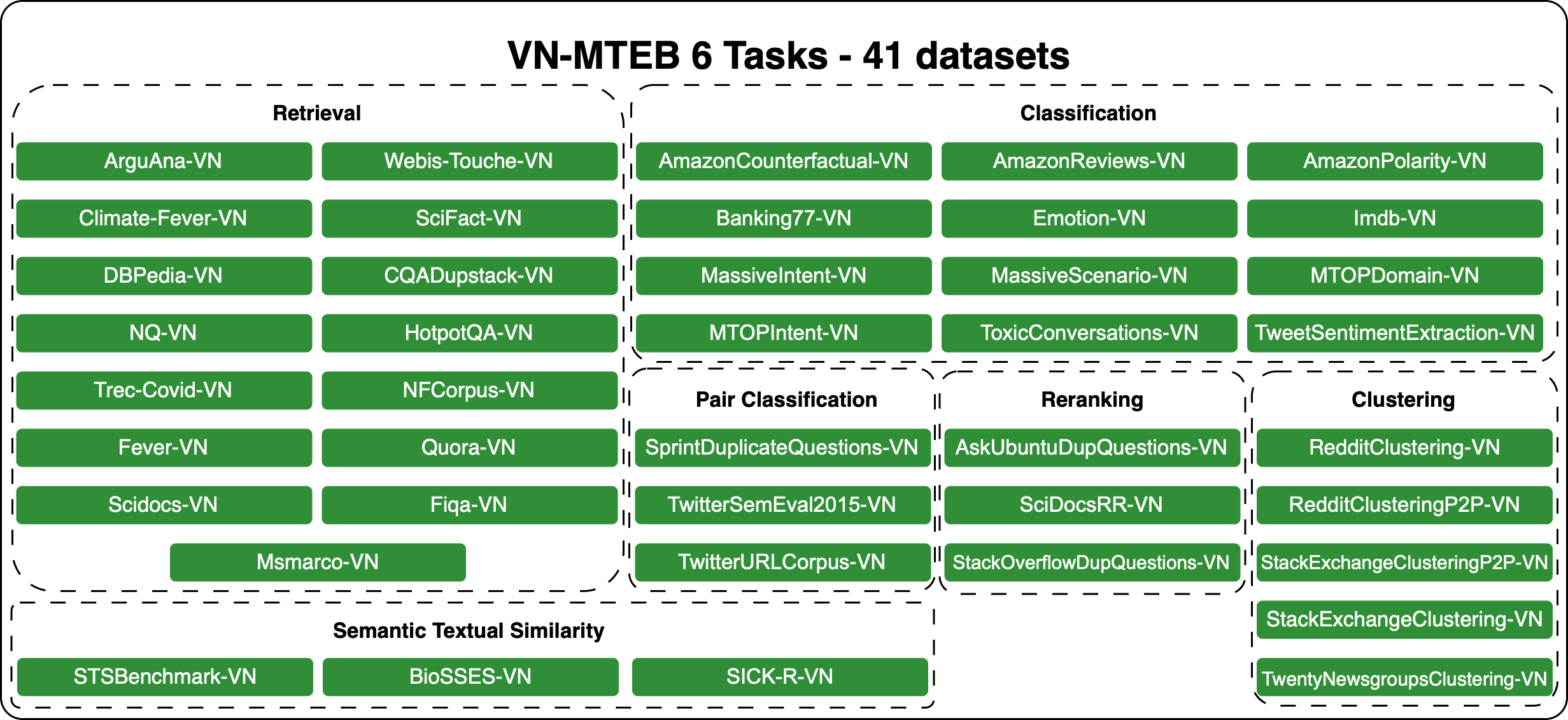}
  \caption{An overview of tasks and datasets in VN-MTEB.}
  \label{fig:vn-mteb}
\end{figure*}
GLUE \cite{wang-etal-2018-glue} and SuperGLUE \cite{wang-etal-2019-superglue}, Big-BENCH \cite{srivastava2023beyond}, and evaluation frameworks \cite{eval-harness} play a crucial role in driving NLP progress. However, they are not suitable for evaluating text embedding, so dedicated benchmarks such as SentEval \cite{conneau-kiela-2018-senteval}, often known as a benchmark for semantic textual similarity (STS), USEB \cite{wang-etal-2021-tsdae-using}, introduced with additional reranking tasks, and Beir \cite{Thakur-2021-BEIR} have become the standard for embedding evaluation for zero-shot information retrieval. The MTEB \cite{muennighoff-etal-2023-mteb} incorporates the above benchmarks and consists of 58 datasets covering 112 languages from 8 embedding tasks: bitext mining, classification, pair classification, clustering, reranking, retrieval, semantic textual similarity (STS), and summarization. Our work follows the structure and is compatible with the current working source of MTEB.

Up until now, the evaluation of text embeddings in the Vietnamese language has primarily focused on individual tasks. The MTEB framework includes some datasets for evaluation, such as VietQuAD2.0 \cite{nguyen-etal-2020-vietnamese-vietquad} for retrieval, VieMedEVBitextMining \cite{vo-etal-2024-improving-med} for bitext mining, and VieStudentFeedbackClassification \cite{uit-vsfc} for classification. Most existing Vietnamese monolingual embedding models are benchmarked on a limited number of individual tasks, such as \texttt{sup-SimCSE-Vietnamese-phobert-base} \footnote{https://huggingface.co/VoVanPhuc/sup-SimCSE-VietNamese-phobert-base} (evaluated on  STSbenchmark dataset \cite{huggingface:dataset:stsb_multi_mt}, STS task), \texttt{vietnamese-bi-encoder} \cite{vietnamese-bi-encoder} and \texttt{Vietnamese-Embedding} \footnote{https://huggingface.co/AITeamVN/Vietnamese\_Embedding} (evaluated on Zalo Legal Text Retrieval dataset \footnote{https://challenge.zalo.ai}, retrieval task). Our VN-MTEB integrates a wide range of datasets, including clustering, classification, BEIR (retrieval) \cite{Thakur-2021-BEIR}, and others from various tasks, to provide a comprehensive and reliable performance assessment of text embedding models in Vietnamese.
\vspace{-4pt}
\subsection{Translation Pipeline}\label{subsec:translation_pipeline}
In Beir-PL \cite{wojtasik-etal-2024-beir-pl}, the verification process involved randomly selecting 100 query-passage pairs, assessed by a linguist in a strict setting and a researcher in a semantic setting. Additionally, an automated comparison was conducted using the multilingual LaBSE model \cite{feng-etal-2022-language-agnostic-bert}, as in the original paper, to compare source texts and translations automatically. The paper applied machine translation with a large language model \cite{yang-etal-2023-human-in-the-loop-machine-translation-with-large-language-model}, where the LLM first generates a draft translation. The pipeline then retrieves similar translation pairs and feedback from the database as in-context examples, allowing the model to refine the draft based on these domain-specific revisions. Furthermore, LLM can be used with various prompt templates to predict human-annotated direct assessment for translation quality \cite{qian-etal-2024-what-do-large-language-models-need-for-machine-translation-evaluation}. They also explored different prompting techniques, including chain-of-thought (CoT) \cite{10.5555/3600270.3602070-chain-of-though}, which involves a two-step process where the LLM first analyzes the differences between the machine translation output and the reference and then scores the translations based on its analysis. In our method, we utilize the embedding model to compare the equivalence between the original text and its translation, while the LLM analyzes and scores the translation quality, allowing us to create a high-quality translated dataset without relying on human effort.

\subsection{Embedding models}\label{subsec:embed_models}
Embedding models create vector representations for tokens, with a key challenge being how they handle positional information in sequences. Our paper extends the foundation laid by \cite{zhu-etal-2024-longembed} on classifying embedding models. It explores architectures like Absolute Positional Encoding (APE) and Rotary Positional Encoding (RoPE), alongside tuning strategies including Instruct-tuned and Non-Instruct-tuned methods. To incorporate positional embeddings into token embeddings, most encoder-based text embedding models, such as the BERT architecture \cite{devlin-etal-2019-bert}, adopt the APE approach. In contrast, the RoPE method \cite{su2023-roformer-enhanced-transformer-rotary} encoded positional information through rotational transformations applied directly to the query and key vectors within the attention mechanism. This approach adopted positional encoding strategies in the age of LLMs, with its use seen in models like LLaMA \cite{touvron2023-llama-open-efficient-foundation} and Qwen \cite{bai2023-qwen-technical-report}.

The Instruct-tuned Model refers to models that were trained with the natural language descriptions of the embedding tasks. Instructions can better inform embedding models about the task at hand, thereby enhancing the quality of the embeddings.

\section{Methodology}
\label{method}
\begin{figure*}[ht!]
  \centering
  \includegraphics[width=0.9\textwidth]{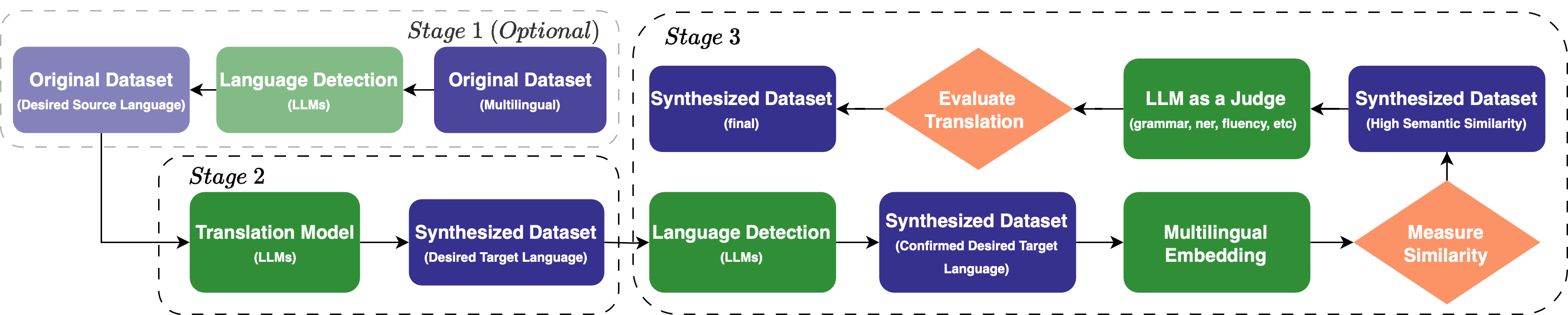}
  \caption{An overview of translation pipeline.}
  \label{fig:translation_pipeline}
\end{figure*}

Our goal is to create a large-scale benchmark that serves as a reference point for comparing different text embedding models in Vietnamese. To achieve this, we focus on a language with a substantial volume of data instances available in the MTEB benchmark and translate its dataset into Vietnamese. For each criterion, we explore the flexible use of embedding models or the application of CoT prompting techniques \cite{10.5555/3600270.3602070-chain-of-though} in large language models to perform evaluation. The objective is to select high-quality synthesized samples while maintaining performance and ensuring resource efficiency.

%
The Figure~\ref{fig:translation_pipeline} illustrates our pipeline for generating a synthesized dataset by transforming a source dataset into a low-resource language. Our pipeline consists of three main stages:
\begin{itemize}
    \item \textbf{Stage 1:} The purpose of this stage is to filter out only the samples in the desired source language. Supposing the original dataset is multilingual, we employ language detection using a LLM to detect the language in the original dataset, keeping only samples in the desired source language. Future studies aiming to translate the entire dataset may omit this stage.
    \item \textbf{Stage 2:} This stage employs the LLM to translate the dataset. The result is a set of Vietnamese sequences that exhibit high similarity to the original texts while preserving semantic fidelity, named entity recognition (NER), code snippets, and other critical aspects, which will be further examined and evaluated in the subsequent stage.
    \item \textbf{Stage 3:} We evaluate the generated translations used in the official VN-MTEB through a three-step process, with each step reflecting an increasing level of rigor. First, we assess whether the data contains any contamination from other languages. Second, we ensure that the data preserves high semantic similarity with the original content. Finally, we score each synthesized sample based on a combination of multiple evaluation criteria. We discard all data samples whose scores fall below the predefined threshold.
\end{itemize}

\textbf{Translation.}
The generated sequences must achieve high quality to minimize the likelihood of being filtered out during the validation stage. Therefore, selecting an appropriate LLM is crucial. In this stage, we recommend using an LLM with at least a medium-sized model and support for maximum token lengths in the tens of thousands. Additionally, we consider utilizing models that demonstrate strong performance on the target language by consulting relevant leaderboards, such as SEA-HELM\footnote{https://leaderboard.sea-lion.ai/}.

Evaluating the quality of model-generated translations is crucial, as embedding models require high-quality datasets for both training and testing. Therefore, we propose a series of data filtering steps to ensure that the final synthesized dataset preserves essential NLP properties while optimizing the framework's execution efficiency. 

\textbf{Language Detection.} We employ a lightweight LLM for language detection to identify samples in the desired source language for translation (Stage 1). While LLMs are generally proficient at translating text, they may misidentify the language when multiple languages are present or when the text includes uncommon phrases, regional dialects, or jargon \cite{qian-etal-2024-what-do-large-language-models-need-for-machine-translation-evaluation}. Additionally, translations may not always capture contextual nuances, idioms, or cultural subtleties. In \cite{qian-etal-2024-what-do-large-language-models-need-for-machine-translation-evaluation}, the shortcomings noted in the LLM's initial translation output are primarily related to domain-specific nuances, terminology, and sometimes word order or structure. Therefore, we also leverage the same language detection model used in Stage 1 to verify whether the translated outputs are entirely in Vietnamese in Stage 3.


\textbf{Semantic Similarity.} The translated text must maintain semantic equivalence with the original sentence. Therefore, we consider using multilingual embeddings to compute similarity scores between sentence pairs and subsequently filter the data based on a predefined threshold. A key factor in selecting an evaluation model is ensuring that the inferred score distributions for similar and unrelated sentence pairs are well separated. Additionally, the model's maximum sequence length should be relatively large (preferably greater than or equal to 8192 tokens) to fully encode the content of each sequence. To determine the optimal threshold for specific models, we need to balance the separation of similarity scores between semantically related and contradictory pairs while minimizing the number of incorrectly filtered samples. (See Section~\ref{sec:experiments} for a more detailed discussion.). 

\textbf{LLM as a Judge.} 
In addition to ensuring consistency in the target language and maintaining semantic similarity to the input sequence, other criteria should also be considered to guarantee that the synthesized samples are of high quality and aligned with human knowledge. Since translation is fundamentally about generating text that is both accurate and aligned with human linguistic expectations in a different language, the findings of \cite{10.5555/3666122.3668142-judging-LLM-as-a-judge} are directly relevant to and encouraging for the application of LLM-as-a-Judge for quality assurance in LLM-based translation. The advantages discussed in the paper include scalability and explainability, which support the reason why we are using LLM to judge a large-scale dataset's translation quality. In this paper, we leverage LLMs at this stage to evaluate the following criteria: grammar, named entity recognition (NER), numbers/links/special characters, fluency, and meaning preservation. The following generalized formula computes the final score for each output:
\vspace*{-5pt}
\begin{align}
    \text{score}_{\text{LLM\_judge}} = \frac{\sum\limits_{i \in S} \alpha_i \cdot \text{score}_i}{|S|},
\end{align}
\vspace*{-2pt}
where \( S \) is the set of evaluation criteria, \( \sum_{i \in S} \alpha_i = 1\), \( \alpha_i \) and \( \text{score}_i \in [1, 5] \) denote the importance weight and the score of criterion \( i \), respectively. Synthesized translations whose score $score_{LLM\_judge}$ exceeds the threshold $\xi_{LLM\_judge}$ are selected.
\vspace*{-5pt}
\section{VN-MTEB}
\begin{figure}[H]
    \centering
    \includegraphics[width=1.0\linewidth]{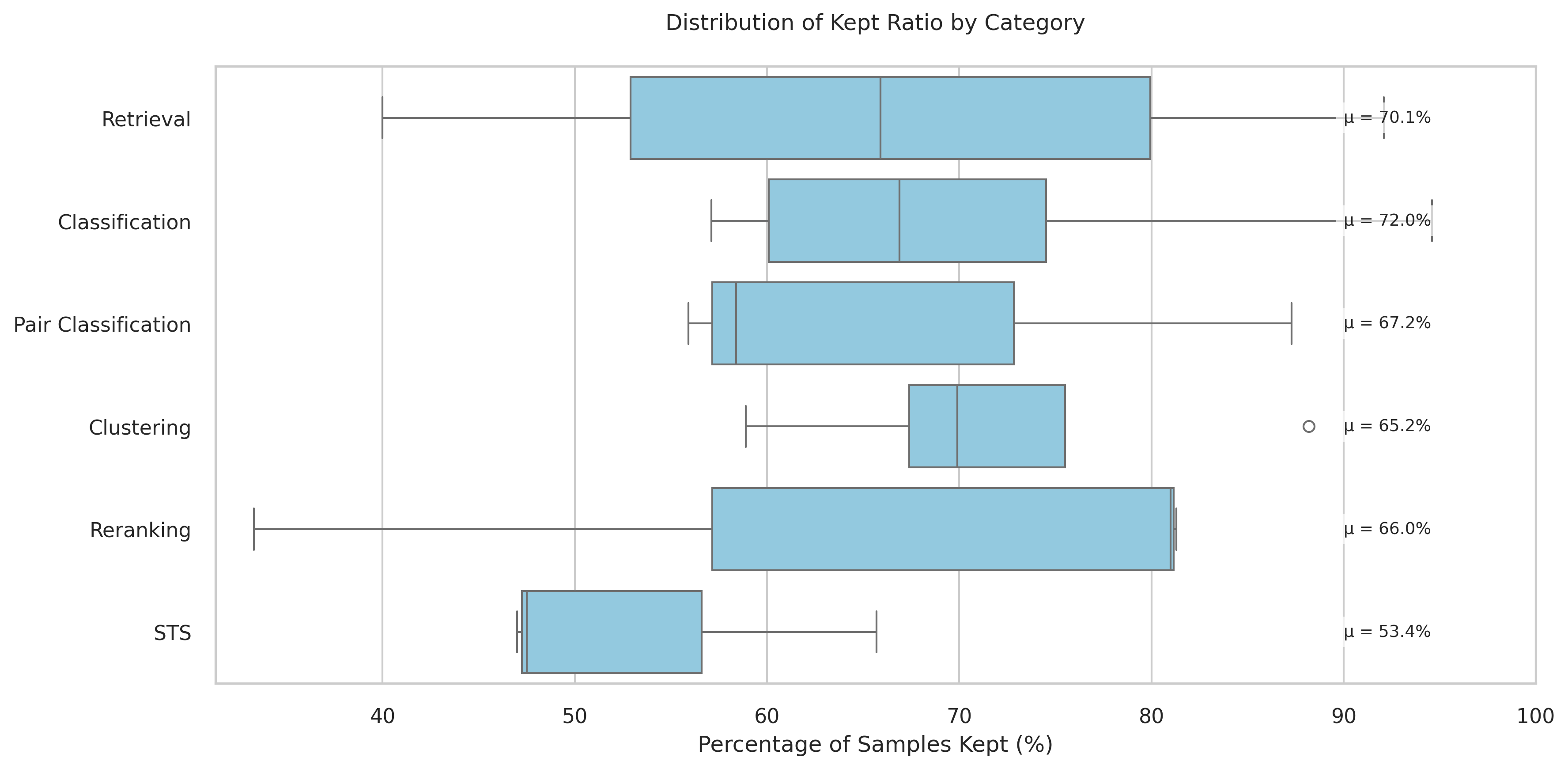}
    \caption{Kept Ratio by Tasks.}
    \label{figure:kept-ratio-by task}
\end{figure}
In Table \ref{tab:dataset_overview}, we provide an overview of the sample collection and count from the original dataset (labeled as "Before") and the final samples obtained after processing through the translation pipeline. For certain tasks, such as re-ranking, clustering, and pair classification, the dataset structure is not strictly sequence-to-sequence. Instead, we structure it as either a sequence or a list of sequences within another list. In our approach, we treat each sequence as an individual sample for the purpose of Stage 3, which is translation validation. Consequently, the sample count may differ from that of the original dataset \cite{muennighoff-etal-2023-mteb} and the dataset statistic \ref{appendix:dataset-statistics} after formatting to be compatible with MTEB code. To the best of our knowledge, this is the first research to release large-scale datasets, which cover the diverse set of tasks for benchmarking Vietnamese embedding models, comprising 41 datasets across 6 tasks.

\begin{table}[ht!]
  \caption{The overview of \textbf{VN-MTEB}.}
  \label{tab:dataset_overview}
  \centering
  \small
  \resizebox{0.5\textwidth}{!}{%
    \begin{tabular}{@{}l@{\hskip 5pt}c@{\hskip 5pt}c@{\hskip 5pt}c@{}} 
    \hline
    Dataset & \# Samples & \# Samples & \% Kept \\
    Name & (Before) & (Final) & (Final/Before) \\
    \hline
    \multicolumn{4}{@{}c@{}}{\textbf{Retrieval}} \\
    \texttt{ArguAna-VN} & 1,406  &  1,295  &  92.1\%  \\
    \texttt{Touche2020-VN} &  2,214 &  1,138 &  51.4\% \\
    \texttt{ClimateFEVER-VN} & 4,681  &  3,401  &  72.6\%  \\
    \texttt{CQADupstack-*-Retrieval-VN} &  19,938 &  13,140  &  65.9\% \\
    \texttt{DBPedia-VN} & 49,188  &  39,551 &  80.4\%  \\
    \texttt{FEVER-VN} &  16,016 &  12,739 &  79.5\% \\
    \texttt{FiQA2018-VN} & 1,706  &  1,021  &  59.8\%  \\
    \texttt{HotpotQA-VN} & 25,704  &  21,956  &  85.5\%  \\
    \texttt{MSMARCO-VN} &  16,697 &  8,019  &  48.0\% \\
    \texttt{NFCorpus-VN} &  12,334 &  6,819  &  55.2\%  \\
    \texttt{NQ-VN} & 4,201  &  2,283  &  54.4\% \\
    \texttt{QuoraRetrieval-VN} &  23,301 &  17,135  &  73.5\% \\
    \texttt{SCIDOCS-VN} & 29,928  &  11,969 &  40.0\%  \\
    \texttt{SciFact-VN} & 339  &  155  &  45.7\%  \\
    \texttt{TRECCOVID-VN} & 66,336  &  57,358 &  86.4\%  \\
    \multicolumn{4}{@{}c@{}}{\textbf{Classification}} \\
    \texttt{EmotionVNClassification} & 4,000 & 2,570   &  64.3\%  \\
    \texttt{Banking77VNClassification} & 13,083  &  12,378  &  94.6\%  \\
    \texttt{ToxicConversationsVNClassification} & 50,000  &  28,560  &  57.1\%  \\
    \texttt{ImdbVNClassification} & 25,000  &  22,081  &  88.3\%  \\
    \texttt{TweetSentimentExtractionVNClassification} & 3,534  &  2,065  &  58.5\%  \\
    \texttt{AmazonCounterfactualVNClassification} & 1,005  &  711  &  70.7\%  \\
    \texttt{MTOPDomainVNClassification} & 30,517  &  20,414  &  66.9\%  \\
    \texttt{MTOPIntentVNClassification} & 30,517  &  20,414  &  66.9\%  \\
    \texttt{AmazonReviewsVNClassification} & 9,990  & 6,766   &  67.8\%  \\
    \texttt{MassiveIntentVNClassification} & 5,005  & 3,005   &  60.1\%  \\
    \texttt{MassiveScenarioVNClassification} & 5,006  &  3,006  &  60.1\%  \\
    \texttt{AmazonPolarityVNClassification} & 400,000  & 344,197   &  86.0\%  \\
    \multicolumn{4}{@{}c@{}}{\textbf{Pair Classification}} \\
    \texttt{SprintDuplicateQuestions-VN} & 202,000  &  176,259  &  87.3\%  \\
    \texttt{TwitterSemEval2015-VN} & 16,777  &  9,374  &  55.9\%  \\
    \texttt{TwitterURLCorpus-VN} &  51,534 &  30,111  &  58.4\%  \\
    \multicolumn{4}{@{}c@{}}{\textbf{Clustering}} \\
    \texttt{TwentyNewsgroupsClustering-VN} & 59,436  &  45,034  &  58.9\%  \\
    \texttt{RedditClustering-VN} & 190,653 &  133,217  & 69.9\%   \\
    \texttt{RedditClusteringP2P-VN} & 438,322  & 331,020   &  75.5\%  \\
    \texttt{StackExchangeClustering-VN} & 35,052  & 23,618   &  67,4\%  \\
    \texttt{StackExchangeClusteringP2P-VN} & 73,577  &  64,869  &  88,2\%  \\
    \multicolumn{4}{@{}c@{}}{\textbf{Reranking}} \\
    \texttt{AskUbuntuDupQuestions-VN} & 375  & 305 &  81.3\%  \\
    \texttt{StackOverflowDupQuestions-VN} & 2,992  & 2,421   &  81.0\%  \\
    \texttt{SciDocsRR-VN} & 7,959  &  2,656  &  33.3\%  \\
    \multicolumn{4}{@{}c@{}}{\textbf{Semantic Textual Similarity}} \\
    \texttt{STSBenchmark-VN} &  2,879 &  1,891  &  65.7\%  \\
    \texttt{BIOSSES-VN} & 100  & 47   &  47.0\%  \\
    \texttt{SICK-R-VN} &  9,927 &  4,716  &  47.5\%  \\
  \end{tabular}%
  }
\end{table}

\textbf{Kept ratio.} The percentage of retained samples (\% Kept) is determined by the ratio of the final sample count to the original sample count. The varying kept ratios suggest different levels of data quality and filtering requirements across tasks. Higher kept ratios generally indicate more reliable or cleaner datasets.
Lower kept ratios might indicate either more challenging data domains or stricter task-specific requirements. Some datasets have a kept ratio lower than 50\%, indicating that half of the translations were invalid due to complexities in grammar and semantics, which are difficult to translate, as well as issues with passing quality control in Stage 3 of our pipeline. Further implementation detail please refer to section ~\ref{sec:experiments}.

\begin{figure}[H]
    \centering
    \includegraphics[width=0.8\linewidth]{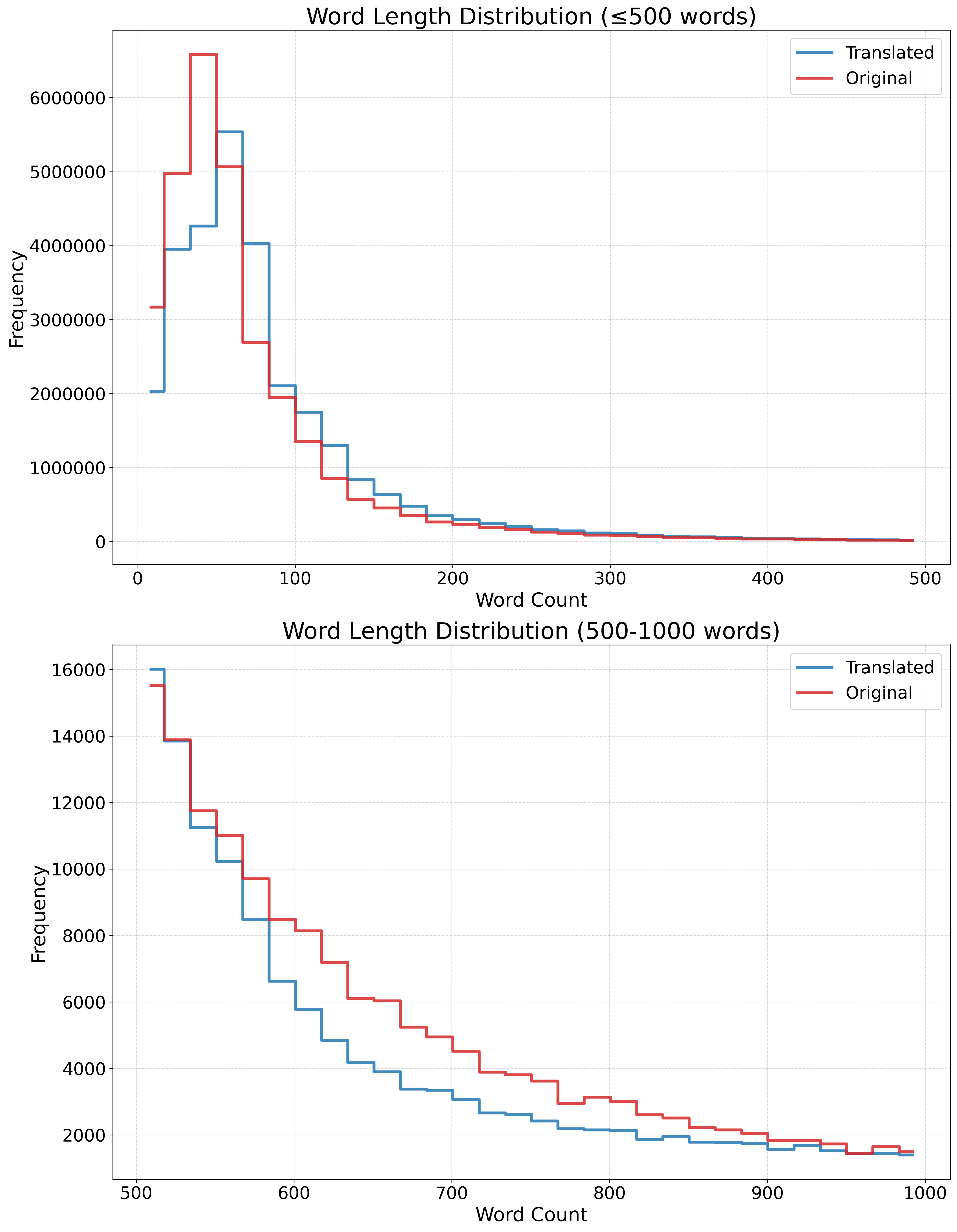}
    \caption{Word Length Distribution between Original and Translated in overall dataset.}
    \label{figure:word_len_distribution}
\end{figure}

\textbf{Word length.} Since both English and Vietnamese originate from Latin roots, analyzing the distribution of word lengths between original and synthesized samples has the potential to reflect translation quality. We conduct a statistical analysis over a word length range that covers the majority of samples in the VN-MTEB dataset. Figure~\ref{figure:word_len_distribution} compares the distributional trends over a dataset consisting of millions of sample pairs. The results reveal a strong correlation between Vietnamese and English word lengths. This observation serves as supporting evidence for translation quality assessment, in addition to the evaluation criteria discussed in Section~\ref{method}.

For more detailed statistics, please refer to our table \ref{tab:appendix-dataset-detail} for information on the train, dev, and test split samples, and see \ref{appendix:gpu-usage} for further details about GPU usage and the time spent creating all datasets.
\section{Experiments}\label{sec:experiments}
\subsection{Implementation Details}
In this part, we provide a detailed report on the models and hyperparameters used for dataset translation and verification. In our pipeline, we refer to the Seahelm leaderboard\footnote{https://leaderboard.sea-lion.ai} and select \texttt{Qwen/Qwen2.5-3B-Instruct} \footnote{https://huggingface.co/Qwen/Qwen2.5-3B-Instruct} to perform detecting language, which was the top model with the relatively small size compared to the time our experiment was conducted. The choice of model at translation stage is guided by a trade-off between translation quality and the computational cost of processing large-scale resources, potentially involving millions of documents. Throughout the course of this research, we evaluated a diverse set of machine translation models, including pre-trained multilingual models such as \texttt{SeamlessM4T} \cite{communication-2023-seamlessmultilingualexpressivestreaming}, \texttt{M2M100} \cite{fan-2020-englishcentricmultilingualmachinetranslation}, and \texttt{NLLB-200} \cite{nllbteam-2022-languageleftbehindscaling}, all of which represent significant advancements in cross-lingual representation learning. Additionally, we considered state-of-the-art bilingual translation models tailored specifically for English–Vietnamese translation, including \texttt{EnViT5-Translation} \cite{ngo2022mtetmultidomaintranslationenglish} and \texttt{VinAI-Translate-En2Vi} \cite{vinaitranslate}. There are limitations of prior machine translation works such as VinAI-Translate-En2Vi \cite{vinaitranslate}, which is short context length (1024) and limitation of domain trained. API-based models like OpenAI's GPT-4, Google's Gemini, etc are costly to translate on a massive dataset. To process and translate the available MTEB benchmark's dataset into Vietnamese, we have done several experiments to identify the best model. At the time the experiment and translation were conducted, we chose the best model according to SouthEast Asian Holistic Evaluation of Language Models (SEA Healms) \footnote{https://leaderboard.sea-lion.ai} that time (May 23, 2024), we used Coherence AI's \texttt{Aya-23-35B} \cite{aryabumi-2024-aya23-openweight}, which has relatively good performance on Vietnamese, and the model size is relatively feasible (35 billion parameters). We utilize the embedding model \texttt{Alibaba-NLP/gte-Qwen2-7B-instruct} \footnote{https://huggingface.co/Alibaba-NLP/gte-Qwen2-7B-instruct}text to compute semantic similarity for embedding-based evaluations. The advantage of deploying this model lies in its ability to encode long sequences (up to 32,768 tokens). For the "LLM-as-a-Judge" evaluation framework, we adopt \texttt{aisingapore/Llama-SEA-LION-v3-70B-IT} as the scoring model. According to the SEA Healms benchmark, this model currently demonstrates the strongest performance for Vietnamese. To enhance judgment quality, we further incorporate chain-of-thought (CoT) prompting techniques in the evaluation process.

In our research, we used 4 NVIDIA H100 GPUs to run our pipeline. For a full estimate about the resource usage, please refer to Appendix GPU usage \ref{appendix:gpu-usage}, and for LLMs hyperparameters in translation, please refer to Appendix table \ref{tab:translation-hyperparams} .

\subsection{Experimental Results}
\textbf{Language Detection.} A conventional approach for language detection on text sequences is to employ FastText \cite{joulin-etal-2017-bag-of-tricks-fasttext}. However, synthesized texts often contain interleaved characters from multiple languages, as discussed in Section~\ref{method}. Through our experiments, we demonstrate that FastText frequently yields inaccurate predictions in such cases. Consequently, leveraging a lightweight large language model (LLM) in conjunction with the CoT technique proves to be a more effective solution for detecting the language of generated samples. Visual results are presented in Table~\ref{tab:lang-identification}.

\begin{table*}[!ht]
\centering
\tiny  
\setlength{\tabcolsep}{8pt} 
\resizebox{0.80\textwidth}{!}{
\begin{tabular}{|c|p{5cm}|c|c|c|c|}
\hline
\textbf{Dataset Name} & \textbf{Translated Text} & \textbf{True Label} & \textbf{ Qwen2.5-7B-Instruct} & \textbf{ Qwen2.5-3B-Instruct} & \textbf{ FastText} \\ 
\hline
cqadupstack-mathematica-vn & Dựa trên một tập dữ liệu, có cách nào để thay đổi một giá trị? Ví dụ (\textit{data = First@Import[``dataset.xlsx''];}) data= \{\{``Supplier'', ``Material'', ``Geography'', ``Quantity''\}, \{``Acme'', ``A'', ``United States'', 676.\}\ldots & vie\_Latn & vie\_Latn & vie\_Latn & \textcolor{red}{krc\_Cyrl} \\
webis-touche2020-vn & 2007 Hall of Fame BBWAA \textit{(98,5\%) Được chọn vào HOF năm 2007 bởi BBWAA All-Star Games 1983 * 1984 (SS) 1985 (SS) 1986 (SS) 1987 (SS) 1988 (SS) 1989 (SS) 1990}\ldots & vie\_Latn & vie\_Latn & vie\_Latn & \textcolor{red}{kor\_Hang} \\
msmarco-vn & Ga Amtrak gần Buena Park: 1  5 dặm: FULLERTON (120 E. SANTA FE AVE.) . 2  8 dặm: ANAHEIM (2150 KATELLA AVE.) . 3  12 dặm: SANTA ANA (1000 E. SANTA ANA BLVD.)\ldots & vie\_Latn & vie\_Latn & vie\_Latn & \textcolor{red}{kor\_Hang} \\
\hline
\end{tabular}}
\caption{Comparison of Vietnamese Language Identification: Qwen2.5-7B-Instruct vs Qwen2.5-3B-Instruct vs. FastText.}
\label{tab:lang-identification}
\end{table*}

\textbf{Translation.} Table~\ref{tab:dataset_overview} presents the results obtained using the selected translation model \texttt{Aya-23-35B} \cite{aryabumi-2024-aya23-openweight}. Our pipeline demonstrates strong translation performance across most datasets, achieving a relatively high retention rate and satisfactory quality in terms of preserving semantic meaning, named entities, and other key elements. Although some datasets, such as \texttt{SciDocsRR-VN}, \texttt{SCIDOCS-VN}, and \texttt{Scifact-VN}, exhibit retention rates below 50\%, these belong to the scientific domain, which poses particular challenges for translation. 

\begin{figure}
    \centering
    \includegraphics[width=0.8\linewidth]{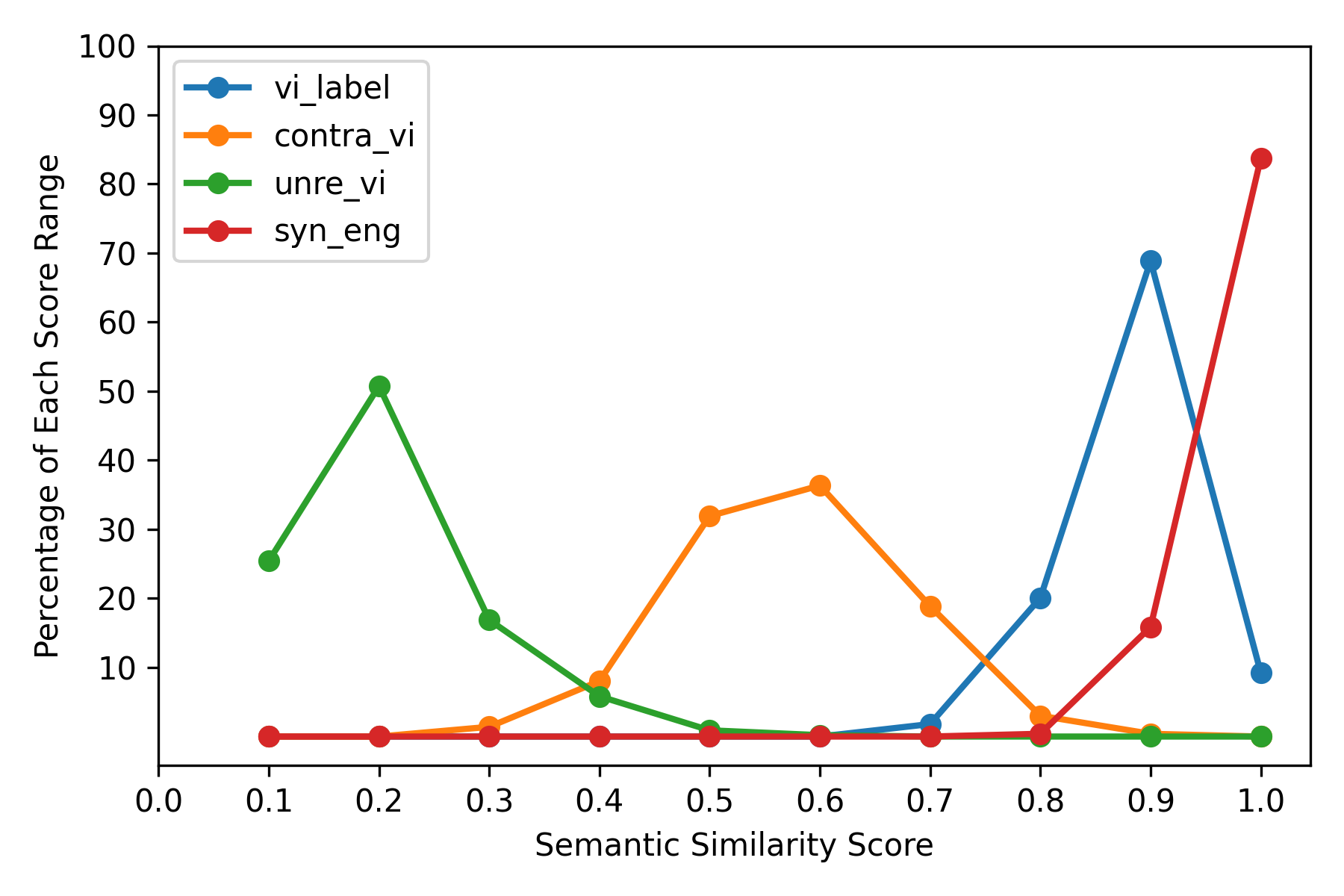}
    \caption{The distribution of semantic similarity score using \texttt{Alibaba-NLP/gte-Qwen2-7B-instruct}. vi\_label, contra\_vi, unre\_vi, and syn\_eng respectively represent the semantic similarity scores between the original English sequences and the corresponding labeled Vietnamese sequences, contrastive Vietnamese sequences, unrelated Vietnamese sequences, and synonymous English sequences.}
    \label{figure:semantic_score}
\end{figure}

\textbf{Semantic Similarity.}  Figure~\ref{figure:semantic_score} illustrates the percentage distribution of semantic similarity score regions (binned in intervals of 0.1) for different sentence pairs, including original English sentences with their corresponding Vietnamese labels, semantically similar English sentences, contradictory Vietnamese sentences, and unrelated Vietnamese sentences. We evaluate 500 samples from the FLoRes \footnote{https://github.com/facebookresearch/flores} dataset, which provides pre-aligned English-Vietnamese sentence pairs. The remaining sentence categories for semantic comparison are manually curated by bilingual experts. The results presented in Figure~\ref{figure:semantic_score} indicate a clear separation in the semantic similarity score distribution between original English sentences paired with their Vietnamese labels and semantically similar English sentences, compared to the other sentence pairs. Based on these results, we discard generated texts that scores fail to satisfy the minimum threshold of 0.8. 

\begin{figure}
    \centering
    \includegraphics[width=0.9\linewidth]{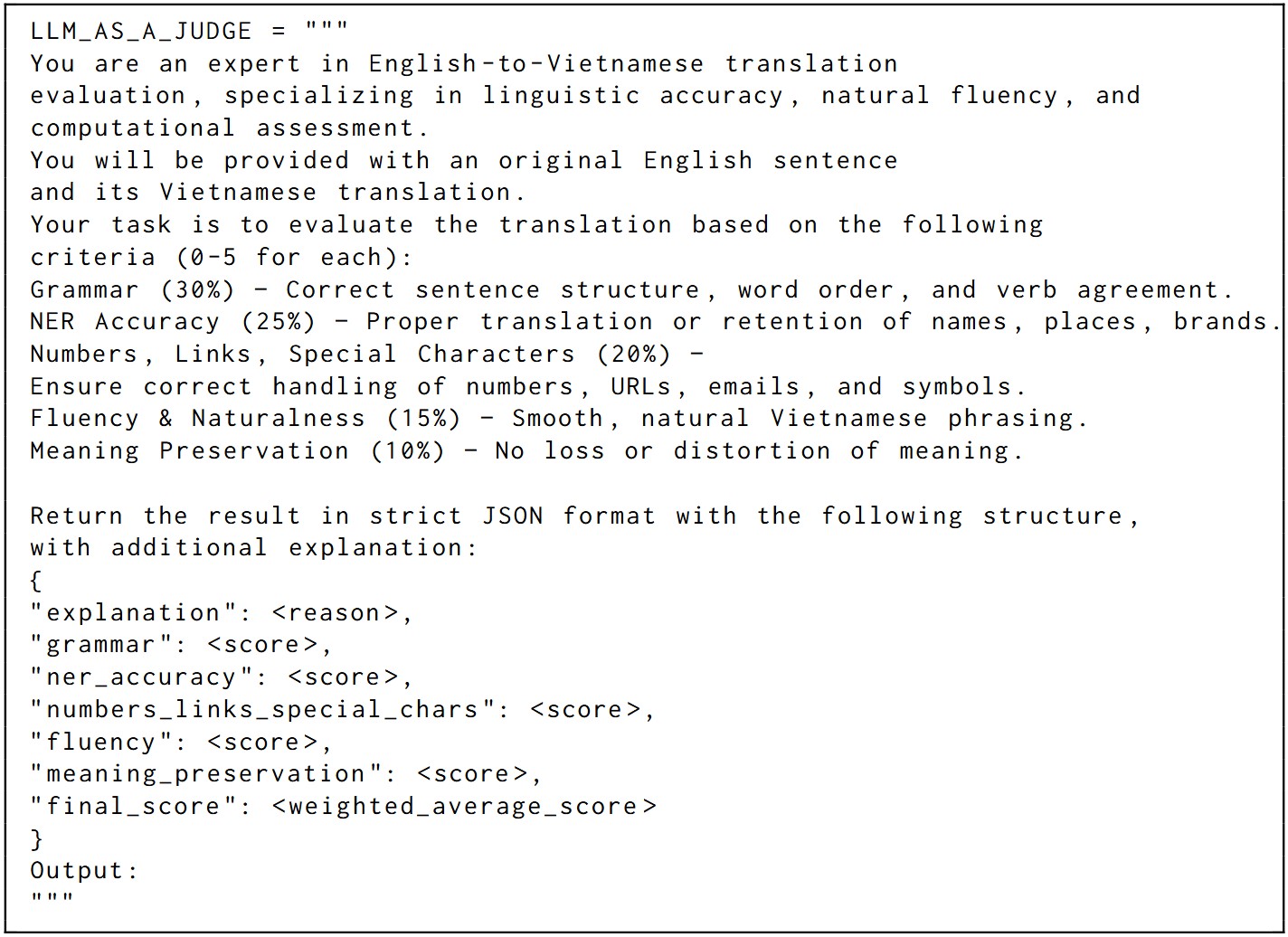}
    \caption{LLM as a Judge prompt.}
    \label{figure:llm-as-a-judge-prompt}
\end{figure}

\textbf{LLM as a Judge.} This step involves evaluating translations based on criteria such as grammar, named entities, fluency, and more. Since translation is essentially about producing text that is both accurate and conforms to human linguistic standards in another language, the findings from \cite{10.5555/3666122.3668142-judging-LLM-as-a-judge} are relevant and encouraging for using LLM-as-a-Judge in quality assurance for LLM-based translations. The paper highlights advantages such as scalability and explainability, which justify using LLM to assess translation quality across large datasets. Although the LLM as a Judge has limited reasoning, with Chain-of-Thought (CoT) prompting techniques \cite{10.5555/3600270.3602070-chain-of-though},  CoT guides LLMs in evaluation tasks by breaking down the entire evaluation process into smaller steps with detailed definitions and constraints for each step in the prompts. We used this technique to design the prompt guiding the LLM to step-by-step generate an explanation and then scoring the translation.
We're using a prompt that is described in Figure~\ref{figure:llm-as-a-judge-prompt}. 

\begin{table*}[ht!]
    \centering
    \resizebox{0.80\textwidth}{!}{\begin{tabular}{l|ccc|cccccc|c}
    \hline
 & Size & Dim & Type & Retr. & Class. & PairClass. & Clust. & Rerank. & STS & Avg. \(\uparrow\) \\
Num. Datasets ($\rightarrow$) & (Params) & (Dim) &  & 15 & 12 & 3 & 5 & 3 & 3 & 41 \\
\hline
\texttt{gte-Qwen2-7B-instruct*} & 7B & 3584 & \texttt{RoPE}  & \textbf{46.05} & 70.76 & 72.09 & \textbf{53.15} & 74.28 & 78.73 & 65.84 \\
\texttt{e5-Mistral-7B-instruct*} & 7B & 4096 & \texttt{RoPE} & 41.73 & 72.21 & 84.01 & 51.71 & \textbf{75.15} & 81.20 & 67.67 \\
\texttt{bge-multilingual-Gemma2*} & 9B & 3584 & \texttt{RoPE} & 20.52 & 71.78 & 66.97 & 40.13 & 64.21 & 66.11 & 54.95 \\
\texttt{gte-Qwen2-1.5B-instruct*} & 1.5B & 1536 & \texttt{RoPE} & 42.01 & 67.14 & 72.70 & 47.64 & 71.37 & 79.97 & 63.47 \\
\texttt{m-e5-large-instruct*} & 560M & 1024 & \texttt{APE} & 40.88 & \textbf{73.39} & \textbf{84.47} & 52.96 & 73.28 & \textbf{82.94} & \textbf{67.99} \\
\texttt{m-e5-large} & 560M & 1024 & \texttt{APE} & 37.65 & 65.03 & 83.70 & 45.78 & 70.40 & 80.65 & 63.87 \\
\texttt{bge-m3} & 568M & 1024 & \texttt{APE} & 39.84 & 69.09 & 84.43 & 45.90 & 71.28 & 78.84 & 64.90 \\
\texttt{Vietnamese-Embebedding} & 568M & 1024 & \texttt{APE} & 34.18 & 69.06 & 82.84 & 45.61 & 70.89 & 77.48 & 63.34 \\
\texttt{KaLM-embedding-m-mini-v1} & 494M & 896 & \texttt{RoPE} & 35.07 & 62.84 & 79.95 & 46.85 & 68.85 & 78.54 & 62.02 \\
\texttt{LaBSE} & 471M & 768 & \texttt{APE} & 17.77 & 60.93 & 77.57 & 34.59 & 65.65 & 72.04 & 54.76 \\
\texttt{gte-multilingual-base} & 305M & 768 & \texttt{APE} & 38.38 & 64.99 & 84.42 & 50.25 & 71.78 & 81.51 & 65.22 \\
\texttt{m-e5-base} & 278M & 768 & \texttt{APE} & 34.50 & 63.29 & 82.51 & 45.70 & 69.07 & 79.45 & 62.42 \\
\texttt{halong-embedding} & 278M & 768 & \texttt{APE} & 34.45 & 63.33 & 81.20 & 43.42 & 69.83 & 77.39 & 61.60 \\
\texttt{m-e5-small} & 118M & 384 & \texttt{APE} & 34.12 & 60.27 & 81.18 & 43.16 & 67.69 & 77.56 & 60.66 \\
\texttt{vietnamese-bi-encoder} & 135M & 768 & \texttt{APE} & 25.37 & 58.92 & 77.40 & 34.13 & 64.95 & 68.58 & 54.89 \\
\texttt{sup-SimCSE-VN-phobert-base} & 135M & 768 & \texttt{APE} & 12.03 & 59.69 & 71.31 & 33.05 & 58.86 & 68.61 & 50.59 \\
\texttt{MiniLM-L12} & 33.4M & 384 & \texttt{APE} & 14.14 & 45.57 & 69.46 & 24.36 & 60.44 & 62.34 & 46.05 \\
\texttt{MiniLM-L6} & 22.7M & 384 & \texttt{APE} & 9.65 & 45.19 & 66.13 & 20.40 & 59.46 & 58.25 & 43.18 \\
\hline
\end{tabular}}
\caption{Average performance of the main metric (in percentage) per task and per model on VN-MTEB subsets. The symbol \texttt{*} indicates that the model is \textbf{Instruct-tuned}. Bold values highlight the best results for each specific task. The column "Avg." represents the mean of the average scores across all tasks.}
\label{tab:results-avg}
\end{table*}


The VN-MTEB dataset is the result of considerable efforts in translation and evaluation. Given the constraints of time and resources, we opted to outsource the scoring of translation samples to a large language model (LLM). 

An overview of the final dataset, along with the corresponding Kept ratio, is presented in Table \ref{tab:dataset_overview}.
The mean Kept ratio for the various tasks is as follows: Retrieval (15 datasets) – 66.03\%, Classification (13 datasets) – 70.11\%, Pair Classification (3 datasets) – 67.2\%, Clustering (5 datasets) – 71.98\%, Re-ranking (3 datasets) – 65.2\%, and Semantic Textual Similarity (3 datasets) – 53.4\%.

\subsection{Benchmark Result}
In this paper, we select open-source embedding models to perform benchmarking. In our benchmark, we classified two types of models: APE-based, RoPE-based, and Instruct-tuned models.
Our benchmark results collected from 18 models and averaged from 41 datasets from 6 tasks are represented in Table \ref{tab:results-avg}. For more detail of model scoring on each dataset, please refer to Appendix \ref{appendix:detail-model-result} for results on all of the models we experimented with.

\textbf{Comparison of models}: As visualized in Figure \ref{fig:model-performance-and-size}, there is a clear correlation between the number of parameters in a model and its overall average VN-MTEB score. Larger models tend to achieve higher scores. Specifically, RoPE-based models, such as \texttt{e5-Mistral-7B-Instruct} and \texttt{e5-Qwen2-7B-Instruct}, generally outperform APE-based models like \texttt{gte-multilingual-base}, \texttt{bge-m3}, and \texttt{m-e5-large}. As mentioned in the preliminary section \ref{sec:related_works}, instruct-tuned models were trained with task descriptions. This training approach typically results in higher overall performance, as evidenced by the significant performance improvement of the instruct-tuned \texttt{m-e5-large-instruct} compared to its non-instruct counterpart, \texttt{m-e5-large}.
In the model evaluation process, we adhere to the methodology outlined in the MTEB task \cite{muennighoff-etal-2023-mteb}. Specifically, we employ the model to embed both the queries and the corpus documents for the Retrieval task. Cosine similarity is then used to compute the similarity scores between each query and document. Next, we rank the corpus documents for each query based on their respective similarity scores and calculate various evaluation metrics. It is noteworthy that models with higher-dimensional representations tend to yield improved results in the retrieval task.

\section{Conclusion}
We utilize our proposed translation pipeline for translating 41 datasets from 6 tasks to create a massive text embedding benchmark from English to a low-resource language—Vietnamese. 
Through extensive experiments on our translation pipeline, we show that with LLMs we can delegate lots of effort from humans to translate a massive dataset with quality. Additionally, we evaluated 18 text embeddings and revealed the superiority of RoPE-based embedding models over APE-based ones in some tasks, giving an overview of choices to consider when selecting types of models to put in production and further research. 

\section*{Limitations}
\textbf{Language variability} While this pipeline can be applied to any source language and translated into various low-resource languages, further research and analysis are required to determine the most suitable model for translation. In our study, we have selected LLMs and embeddings based on their performance with English and Vietnamese. For application to other languages, additional experiments must be conducted to identify the most appropriate model for each target language. 

\textbf{Cultural context} Although our work comes from machine translation, datasets are still limited about the cultural context of the translation, such as formal, informal, or the specific dialect used. 

\textbf{Absent of re-generation} Our pipeline does not guarantee the retention of all samples, resulting in some datasets being reduced by nearly half. Therefore, future research should consider incorporating a regeneration mechanism after the evaluation stage to improve the kept ratio.

\textbf{Long context}
The VN-MTEB dataset encompasses a range of text lengths, including sequence-to-sequence, sequence-to-paragraph, and paragraph-to-paragraph formats. However, it lacks datasets comprising very long documents.

\bibliography{anthology,custom}


\appendix 
\label{sec:appendix}

\section{Hyperparameters for Translation}
\raggedright
In our translation pipeline, we used this configuration,



\begin{table}[h!]
\centering
\caption{Translation Hyperparameters}
\resizebox{0.7\linewidth}{!}{ 
\begin{tabular}{ll}
\toprule
\textbf{Hyperparameter} & \textbf{Value} \\
\midrule
temperature & 0.0 \\
max\_new\_tokens & 4096 \\
tensor\_parallel\_size & 4 \\
max\_model\_len & 8192 \\
max\_num\_seqs & 256 \\
vllm\_gpu\_memory\_utilization & 0.95 \\
\bottomrule
\end{tabular}
}
\label{tab:translation-hyperparams}
\end{table}

\section{Examples}
Tables \ref{tab:retrieval_examples}-\ref{tab:sts_examples} provide examples for each dataset for each task. 
\begin{table*}[t!]
    \small
    \resizebox{\textwidth}{!}{\begin{tabular}{l | l | l }
        \toprule
         \multicolumn{1}{l|}{\textbf{Dataset}}    &
        \multicolumn{1}{c}{\textbf{Query}}   &
        \multicolumn{1}{|c}{\textbf{Relevant-Document}} \\
        \midrule
   ArguAna-VN & \multicolumn{1}{p{6cm}|}{Trong mắt công chúng, chính phủ dường như nghi ngờ tất cả mọi người.} & \multicolumn{1}{p{13cm}}{\textit{<Title>} Nhà triết học chính trị cho rằng các quyền dân sự nên bị hy sinh \textit{<Paragraph>} Đây chỉ là một cuộc điều tra như bất kỳ cuộc điều tra nào khác. Chính phủ rõ ràng phải có cách tiếp cận rộng rãi bởi vì bất kỳ lỗ hổng nào cũng có thể bị lợi dụng bởi những kẻ khủng bố vô đạo đức. Đó là một sự cần thiết, mặc dù cùng với những hậu quả không may, nhưng vẫn là sự cần thiết. Còn về đàm phán với những kẻ khủng bố, theo quan điểm của đề xuất này thì lựa chọn này không tồn tại khi đối phó với những kẻ khủng bố có nền tảng chủ nghĩa nguyên lý, vốn theo định nghĩa là không sẵn lòng thỏa hiệp và do đó không thể đàm phán được...} \\ \midrule
   ClimateFEVER-VN & \multicolumn{1}{p{6cm}|}{"Nếu bạn loại bỏ băng giá, có tiềm năng không chỉ là sự bất ổn định của vách băng sẽ bắt đầu xảy ra, nhưng một quá trình được gọi là sự bất ổn định của tấm băng biển", Matthew Wise, một nhà khoa học cực địa tại Đại học Cambridge nói.} & \multicolumn{1}{p{13cm}}{\textit{<Title>}Nam Cực \textit{<Paragraph>} Nam Cực là lục địa phía Nam nhất trên Trái Đất. Nó bao gồm cực Nam Địa lý và nằm ở vùng Nam Cực của Bán cầu Nam, hầu hết về phía nam của Vòng Bắc Cực, và được bao quanh bởi Đại Dương Nam Cực. Với diện tích $14000000 km^2$, đây là lục địa lớn thứ năm trên thế giới. So sánh với Úc thì diện tích của nó gấp đôi nước Úc . Khoảng 98\% lãnh thổ bị băng tuyết che phủ với độ dày trung bình 1,9 km, kéo dài từ những nơi xa nhất về phía bắc đến Bán đảo Tây Nam Cực...} \\ \midrule
   CQADupstack-*-Retrieval-VN & \multicolumn{1}{p{6cm}|}{Làm thế nào để tôi có thể sử dụng Mathematica để tạo ra mã Fortran tốt hơn?} & \multicolumn{1}{p{13cm}}{\textit{<Title>}Tạo mã C/Java hiệu quả giảm thiểu các phép toán \textit{<Paragraph>} Có thể dùng Mathematica để tạo ra mã C/Java nhằm tối thiểu hóa số lượng phép toán thực hiện không? Ví dụ, đối với ma trận nghịch đảo hay định thức? Với biến lưu trữ tốt?} \\ \midrule
   DBPedia-VN & \multicolumn{1}{p{6cm}|}{American sinh đôi nổi tiếng là vận động viên quần vợt chuyên nghiệp người Mỹ} & \multicolumn{1}{p{13cm}}{\textit{<Title>}Giải quần vợt chuyên nghiệp Nam Natomas \textit{<Paragraph>} Giải quần vợt chuyên nghiệp Nam Natomas là một giải đấu quần vợt được tổ chức tại Sacramento, California, Hoa Kỳ từ năm 2005. Sự kiện này là một phần của ATP Challenger Tour và được chơi trên sân cứng ngoài trời.} \\ \midrule
   FEVER-VN & \multicolumn{1}{p{6cm}|}{Bee Gees đã viết ba bài hát cho các nghệ sĩ khác.} & \multicolumn{1}{p{13cm}}{\textit{<Title>} Bee Gees \textit{<Paragraph>} Bee Gees là một nhóm nhạc pop được thành lập vào năm 1958. Thành viên của họ bao gồm ba anh em Barry, Robin và Maurice Gibb. Nhóm đã có những thành công lớn trong nhiều thập niên thu âm nhạc, nhưng họ cũng có hai giai đoạn đặc biệt nổi bật; đó là thời kỳ ca khúc tại vị trí số một trên bảng xếp hạng cuối thập niên 60 và đầu thập niên 70...} \\ \midrule
   FiQA2018-VN & \multicolumn{1}{p{6cm}|}{Các hình thức thay thế cho lương của nhân viên} & \multicolumn{1}{p{13cm}}{\textit{<Paragraph>} Có một vài sáng kiến tiền tệ địa phương ở danh sách Mỹ ở đây. Hầu hết là những nỗ lực để chuẩn bị một giá trị như một mức lương sống, hoặc khuyến khích mạng lưới tiêu thụ địa phương. Nếu bạn ở trong khu vực thu hút của một trong những điều này, hãy xem nếu bạn có thể có được một khoản trợ cấp hoặc vay để bắt đầu (nếu bạn sẵn sàng mua vào triết lý của nhóm như là một mức lương  \$10 tối thiểu)} \\ \midrule
   HotpotQA-VN & \multicolumn{1}{p{6cm}|}{Năm nào thì phim hoạt hình Barbie Thumbelina và Barbie and the Three Musketeers được phát hành?} & \multicolumn{1}{p{13cm}}{\textit{<Title>} Barbie Thumbelina \textit{<Paragraph>} Barbie Thumbelina, hay còn gọi là \"Barbie Presents: Thumbelina\", là một bộ phim Barbie năm 2009 do Conrad Helten và Nishpeksh Mehra đạo diễn. Đây là tập thứ 15 trong loạt phim hoạt hình của Barbie, với sự lồng tiếng của Kelly Sheridan cho nhân vật chính Barbie. Tên gọi của câu chuyện giống như truyện cổ tích \"Thumbelina\" (Cô bé ngón tay) của Hans Christian Andersen nhưng nội dung lại khác nhau.} \\ \midrule
   MSMARCO-VN & \multicolumn{1}{p{6cm}|}{chuyển oz sang gallon} & \multicolumn{1}{p{13cm}}{\textit{<Paragraph>} Có 0.007812500004244 gallon trong một ounce. Một Ounces bằng 0, 078125 Gallon. Định nghĩa của Ounces . Được biết đến với tên gọi là US fluid ounce, đơn vị thể tích cho các chất lỏng được sử dụng như ounce ở Mỹ và các nước khác thực hành hệ thống US Customary.} \\ \midrule
   NFCorpus-VN & \multicolumn{1}{p{6cm}|}{Chất béo bão hòa} & \multicolumn{1}{p{13cm}}{\textit{<Title>} LDL và HDL cholesterol và nồng độ LDL oxy hóa thay đổi ở người bình thường và tăng cholesterol sau khi sử dụng các mức khác nhau của canxi  \textit{<Paragraph>} Bột ca cao giàu polyphenols như catechin và procyanidins, đã được chứng minh trong nhiều nghiên cứu trên động vật về tác dụng ức chế LDL oxy hóa và tạo mảng xơ vữa. Nghiên cứu của chúng tôi đánh giá nồng độ LDL và LDL oxy hóa trong huyết thanh sau khi dùng các lượng khác nhau của bột ca cao (13, 19,5 và 26 g/ngày) ở những người bình thường và tăng nhẹ cholesterol. Trong nghiên cứu so sánh này...} \\ \midrule
   NQ-VN & \multicolumn{1}{p{6cm}|}{phim Silver Linings Playbook được quay ở đâu? } & \multicolumn{1}{p{13cm}}{\textit{<Title>} Silver Linings Playbook\textit{<Paragraph>} Những địa điểm là Upper Darby, Ridley Park và Lansdowne, những cộng đồng nhỏ nằm ngay bên ngoài Philadelphia, Pennsylvania. Mặc dù không được nhắc tên trong phim, nhưng Ridley Park đã được ghi chú ở cuối, và một cảnh sát viên có thể được nhìn thấy đang đeo chữ viết tắt \"RPPD\" trên cổ áo của mình.} \\ \midrule
   QuoraRetrieval-VN & \multicolumn{1}{p{6cm}|}{Những ý tưởng kinh doanh tốt với mức đầu tư thấp ở Ấn Độ là gì?} & \multicolumn{1}{p{13cm}}{\textit{<Paragraph>} Những ý tưởng kinh doanh nhỏ là gì?} \\ \midrule
   SCIDOCS-VN & \multicolumn{1}{p{6cm}|}{Một Phương pháp hai bước để phân cụm dữ liệu hỗn hợp với các thể loại và số học } & \multicolumn{1}{p{13cm}}{\textit{<Title>} Forensics mạng WhatsApp: Giải mã và hiểu các thông điệp tín hiệu cuộc gọi WhatsApp \textit{<Paragraph>} WhatsApp là một ứng dụng nhắn tin di động phổ biến với hơn 800 triệu người dùng. Gần đây, một tính năng gọi điện thoại đã được thêm vào ứng dụng và chưa có phân tích kỹ thuật số toàn diện nào được thực hiện về tính năng này vào thời điểm viết bài báo này. Trong tác phẩm này, chúng tôi mô tả cách chúng tôi có thể giải mã lưu lượng mạng và thu thập các bằng chứng pháp y liên quan đến tính năng gọi điện thoại mới này bao gồm: a) Số điện thoại WhatsApp, b) địa chỉ IP máy chủ WhatsApp, c) mã hóa âm thanh WhatsApp (Opus), d) thời gian gọi điện thoại WhatsApp và e) chấm dứt cuộc gọi điện thoại WhatsApp. Chúng tôi giải thích các phương pháp và công cụ sử dụng để giải mã lưu lượng truy cập cũng như trình bày chi tiết các phát hiện của chúng tôi liên quan đến các thông điệp điều khiển WhatsApp. Hơn nữa, chúng tôi cũng cung cấp cho cộng đồng một công cụ giúp hình dung các thông điệp giao thức WhatsApp.} \\ \midrule
   SciFact-VN & \multicolumn{1}{p{6cm}|}{Sự kích hoạt NFAT4 đòi hỏi sự di chuyển Ca2+ được trung gian bởi IP3R.} & \multicolumn{1}{p{13cm}}{\textit{<Title>} Điều khiển kích hoạt NFAT isoform và biểu hiện gen phụ thuộc NFAT thông qua hai tín hiệu Ca2+ trong tế bào trùng hợp và phân tách không gian \textit{<Paragraph>} Sự kết hợp kích thích-chuyển tự, liên kết kích thích tại bề mặt tế bào với sự thay đổi biểu hiện gen nhân, được bảo tồn trong tất cả các sinh vật nhân thực. Làm thế nào các yếu tố chuyển tự đồng thời được biểu hiện có liên quan chặt chẽ vẫn chưa rõ ràng. Ở đây, chúng tôi cho thấy hai isoform yếu tố chuyển tự phụ thuộc canxi NFAT1 và NFAT4 đòi hỏi các tín hiệu InsP3 và Ca2+ phân biệt để kích hoạt bền vững về mặt sinh lý. ...} \\ \midrule
   Touche2020-VN & \multicolumn{1}{p{6cm}|}{Khuynh hướng tình dục có được xác định khi sinh ra?} & \multicolumn{1}{p{13cm}}{\textit{<Paragraph>} Khuynh hướng tình dục được xác định khi sinh ra. Làm thế nào? Bạn có thể dễ dàng nhìn thấy một em bé là nam hay nữ bằng cách nhìn bộ phận sinh dục của nó. Bộ phận sinh dục nam là dương vật và bộ phận sinh dục nữ là âm đạo. Đơn giản.} \\ \midrule
   TRECCOVID-VN & \multicolumn{1}{p{6cm}|}{Những chiếc mặt nạ nào là tốt nhất để phòng ngừa nhiễm Covid-19?} & \multicolumn{1}{p{13cm}}{\textit{<Title>} Sự lây lan của virus corona chủng mới (SARS-CoV-2): Mô hình hóa và mô phỏng các chiến lược kiểm soát \textit{<Paragraph>} Bệnh dịch viêm đường hô hấp cấp do virus corona đang lan rộng khắp thế giới và tất cả các hệ thống y tế đều bị quá tải. Virus này được đặt tên là SARS-CoV-2. Trong tình hình này, cần phải đưa ra những quyết định hợp lý về cách chăm sóc bệnh nhân bị COVID-19. Báo cáo tỷ lệ mắc bệnh, các triệu chứng chung và các bộ dụng cụ thử nghiệm sẵn có, các chiến lược kiểm soát khác nhau, mô hình phân ngăn cơ bản và một số nghiên cứu hiện tại về dịch tễ học của bệnh được thảo luận và các mô hình đã công bố trước đó được xem xét. ...} \\ \bottomrule\\
    \end{tabular}}
    \caption{Examples of queries and relevant documents for all datasets included in VN-MTEB. (\textit{<Title>}) and (\textit{<Paragraph>}) are used to distinguish the title separately from the paragraph within a document in the table above. These tokens were not passed to the respective models.}
    \label{tab:retrieval_examples}
\end{table*}

\begin{table*}[!t]
    \centering
    \tiny
    \resizebox{\textwidth}{!}{\begin{tabular}{l|l|l}
    \toprule
\multicolumn{1}{l|}{\textbf{Dataset}} & \multicolumn{1}{c}{\textbf{Text}} & \multicolumn{1}{|l}{\textbf{Label}} \\
\midrule
AmazonCounterfactualVNClassification & \multicolumn{1}{p{10cm}|}{Quintus tiên tri rằng họ sẽ trở thành những vị tử đạo một ngày nào đó, nhưng không phải là ngày hôm đó.} & not-counterfactual \\
\midrule
AmazonPolarityVNClassification & \multicolumn{1}{p{10cm}|}{Chúc mừng năm mới\! Pat yêu quý của tôi có một trong những giọng ca tuyệt vời nhất của thế hệ cô ấy. Tôi đã nghe đĩa CD này trong nhiều NĂM và tôi vẫn YÊU nó. Khi tôi có tâm trạng tốt, nó khiến tôi cảm thấy tốt hơn. Tâm trạng xấu chỉ tan biến như đường trong mưa. Đĩa CD này tràn đầy sự sống. Giọng ca thật tuyệt vời và lời bài hát thật tuyệt vời...} & positive \\
\midrule
AmazonReviewsVNClassification & \multicolumn{1}{p{10cm}|}{Không xứng đáng với giá cả và thiết kế nắp rất tệ. Thiết kế nắp vô cùng kém. Không phù hợp để sử dụng hàng ngày. Nắp đậy quá chặt đến nỗi chúng ta phải vật lộn với chai mỗi ngày để mở nắp. Khi bế em bé trong một tay, việc mở nắp là một cơn ác mộng. Ngoài những tính năng siêu an toàn của nắp, chúng còn rất đắt so với các thương hiệu khác. Hãy tránh xa những sản phẩm này cho đến khi họ cải thiện những vấn đề về nắp. Chúng tôi đã nhiều lần làm tổn thương bản thân khi cố gắng mở nắp vì chúng có những cạnh sắc ở cả cạnh trong và ngoài. Không xứng đáng với giá cả.} & 0 \\
\midrule
Banking77VNClassification & \multicolumn{1}{p{10cm}|}{Làm sao tôi có thể tìm thấy thẻ của mình} & card\textunderscore arrival \\
\midrule
EmotionVNClassification & \multicolumn{1}{p{10cm}|}{Tôi cảm thấy mình vẫn đang nhìn vào một tấm vải vẽ trống hoặc một tờ giấy trắng} & sadness \\
\midrule
ImdbVNClassification & \multicolumn{1}{p{10cm}|}{Tôi yêu khoa học viễn tưởng và sẵn sàng chấp nhận nhiều điều. Phim/phim truyền hình khoa học viễn tưởng thường bị thiếu kinh phí, không được đánh giá cao và hiểu lầm. Tôi đã cố gắng thích điều này, tôi thực sự đã cố gắng, nhưng nó giống như so sánh phim truyền hình khoa học viễn tưởng tốt với Babylon 5 và Star Trek...} & negative \\
\midrule
MassiveIntentVNClassification & \multicolumn{1}{p{10cm}|}{Hãy đánh thức tôi lúc 5 giờ sáng trong tuần này} & alarm\textunderscore set \\
\midrule
MassiveScenarioVNClassification & \multicolumn{1}{p{10cm}|}{Ai là người đang chơi bản nhạc này?} & music \\
\midrule
MTOPDomainVNClassification & \multicolumn{1}{p{10cm}|}{Gọi Nicholas và Natasha} & calling \\
\midrule
MTOPIntentClassification & \multicolumn{1}{p{10cm}|}{Tôi còn những nguyên liệu nào?} & GET\textunderscore INFO\textunderscore RECIPES\\
\midrule
ToxicConversationsVNClassification & \multicolumn{1}{p{10cm}|}{Bingo: Mọi thứ luôn liên quan đến sự tăng trưởng dân số. Nếu chúng ta hạn chế nhập cư, chúng ta sẽ có mức tăng trưởng dân số xấp xỉ KHÔNG. Điều đó thật tuyệt vời cho chất lượng cuộc sống và môi trường!} & not toxic \\
\midrule
TweetSentimentExtractionVNClassification & \multicolumn{1}{p{10cm}|}{Tôi rất thích bài hát Love Story của Taylor Swift} & positive \\
    \bottomrule
    \end{tabular}}
    \caption{Classification examples}
    \label{tab:classification_examples}
\end{table*}
\begin{table*}[!t]
    \centering
    \tiny
    \resizebox{\textwidth}{!}{\begin{tabular}{l|l|l}
    \toprule
\multicolumn{1}{l|}{\textbf{Dataset}} & \multicolumn{1}{c}{\textbf{Text}} & \multicolumn{1}{|l}{\textbf{Cluster}} \\
\midrule
RedditClustering-VN & \multicolumn{1}{p{10cm}|}{Một người Úc đích thực là ai?} & australia.txt \\
\midrule
RedditClusteringP2P-VN & \multicolumn{1}{p{10cm}|}{Những chiến thắng không được ghi lại chính xác Hôm nay tôi đã có $5$ chiến thắng trong chế độ solo, nhưng hồ sơ của tôi lại hiển thị $0$ chiến thắng ở chế độ solo và $5$ chiến thắng ở LTM tôi có thể đảm bảo rằng tôi không chơi LTM và chưa bao giờ chơi chế độ này vì đây là tài khoản mới. Có ai gặp phải vấn đề này không? Tôi chơi trên PC. } & FortNiteBR \\
\midrule
StackExchangeClustering-VN & \multicolumn{1}{p{10cm}|}{Thuật ngữ nào tốt hơn cho "front-end" và "back-end" của cơ sở dữ liệu dành cho người dùng phi kỹ thuật?} & ux.stackexchange.com.txt \\
\midrule
StackExchangeClusteringP2P-VN & \multicolumn{1}{p{10cm}|}{Có ai có ví dụ về Dual Contouring trong C\# không? Tôi đang cố gắng phát triển một phương pháp tạo địa hình sử dụng Perlin. Tôi đã theo dõi rất nhiều hướng dẫn của Minecraft và đã khiến chúng hoạt động. Tôi đã thử nghiệm với MarchingSquares, nhưng tôi không thích nó. Bây giờ, tôi đang cố gắng tạo ra một phương pháp dual contouring và tôi cũng đang cố gắng nắm bắt khái niệm về Octrees. Tôi từng phân đoạn mảng dữ liệu của mình thành những phần nhỏ, nhưng việc thu gọn và tạo một "phần" lớn hoạt động giống như một bộ phân đoạn nhỏ hơn không hiệu quả.Tôi hy vọng ai đó có thể chia sẻ một số mã C\#, tốt nhất là dành cho Unity nhưng bất cứ điều gì để tôi có thể phân tích và hiểu cũng sẽ hữu ích.} & unity \\
\midrule
TwentyNewsgroupsClustering-VN & \multicolumn{1}{p{10cm}|}{Windows 3.1 mới bán với giá $\$35$} & 6 \\ \bottomrule
    \end{tabular}}
    \caption{Clustering examples}
    \label{tab:clustering_examples}
\end{table*}

\begin{table*}[!t]
    \centering
    \tiny
    \resizebox{\textwidth}{!}{\begin{tabular}{l|l|l|l}
    \toprule
\multicolumn{1}{l|}{\textbf{Dataset}} & \multicolumn{1}{c}{\textbf{Sentence 1}} & \multicolumn{1}{|c}{\textbf{Sentence 2}} & \multicolumn{1}{l}{\textbf{Label}}\\
\midrule
SprintDuplicateQuestions-VN & \multicolumn{1}{p{5cm}|}{Tại sao tôi không thể tìm ra cách dễ dàng nào để gửi một hình ảnh có văn bản trên Kyocera DuraCore của tôi?} & \multicolumn{1}{p{5cm}|}{Gửi hoặc nhận hình ảnh có văn bản  Kyocera DuraCore} & 1 \\
\midrule
TwitterSemEval2015-VN & \multicolumn{1}{p{5cm}|}{Kết thúc của phim 8 Mile là phần yêu thích nhất của bộ phim.} & \multicolumn{1}{p{5cm}|}{Đó chỉ là lời bài hát rap trong phim 8 Mile} & 0 \\
\midrule
TwitterURLCorpus-VN & \multicolumn{1}{p{5cm}|}{Làm thế nào những ẩn dụ chúng ta sử dụng để miêu tả sự khám phá ảnh hưởng đến nam và nữ trong lĩnh vực khoa học} & \multicolumn{1}{p{5cm}|}{Những ý tưởng lớn đòi hỏi phải có những nỗ lực to lớn, và cách chúng ta nói về chúng cũng rất quan trọng.} & 0 \\
    \bottomrule
    \end{tabular}}
    \caption{Pair classification examples. Labels are binary.}
    \label{tab:pair_classification_examples}
\end{table*}
\begin{table*}[!t]
    \centering
    \tiny
    \resizebox{\textwidth}{!}{\begin{tabular}{l|l|l|l}
    \toprule
\multicolumn{1}{l|}{\textbf{Dataset}} & \multicolumn{1}{c}{\textbf{Query}} & \multicolumn{1}{|c}{\textbf{Positive}} & \multicolumn{1}{|c}{\textbf{Negative}} \\
\midrule
AskUbuntuDupQuestions-VN & \multicolumn{1}{p{5cm}|}{không thể khởi động từ USB} & \multicolumn{1}{p{5cm}|}{USB cài Windows 7 không khởi động sau khi cài Ubuntu} & \multicolumn{1}{p{5cm}}{không thể khởi động từ liveusb được tạo với pendrivelinux} \\
\midrule
SciDocsRR-VN & \multicolumn{1}{p{5cm}|}{Lý thuyết Lãnh đạo phức tạp: Chuyển đổi phong cách lãnh đạo từ thời kỳ công nghiệp sang kỷ nguyên tri thức} & \multicolumn{1}{p{5cm}|}{Lý thuyết lãnh đạo phức tạp: Một quan điểm tương tác về lãnh đạo trong các hệ thống thích ứng phức tạp.} & \multicolumn{1}{p{5cm}}{MedRec: Sử dụng Blockchain cho Truy cập Dữ liệu Y tế và Quản lý Quyền truy cập} \\
\midrule
StackOverflowDupQuestions-VN & \multicolumn{1}{p{5cm}|}{Sử dụng numpy.genfromtxt để đọc một tệp csv với các chuỗi chứa dấu phẩy } & \multicolumn{1}{p{5cm}|}{numpy genfromtxtpandas đọc csv bỏ qua dấu phẩy $;$ trong dấu ngoặc kép } & \multicolumn{1}{p{5cm}}{Lời bình luận của đối số genfromtxt trong numpy} \\
    \bottomrule
    \end{tabular}}
    \caption{Reranking examples}
    \label{tab:reranking_examples}
\end{table*}
\begin{table*}[!t]
    \centering
    \tiny
    \resizebox{\textwidth}{!}{\begin{tabular}{l|l|l|l}
    \toprule
\multicolumn{1}{l|}{\textbf{Dataset}} & \multicolumn{1}{c}{\textbf{Sentence 1}} & \multicolumn{1}{|c}{\textbf{Sentence 2}} & \multicolumn{1}{|c}{\textbf{Score}} \\
\midrule
BIOSSES-VN & \multicolumn{1}{p{5cm}|}{Mutations của gen KRAS gây ung thư là những đột biến phổ biến trong ung thư.} & \multicolumn{1}{p{5cm}|}{Đáng chú ý, c-Raf gần đây đã được phát hiện là yếu tố thiết yếu cho sự phát triển của NSCLC do K-Ras gây ra.} & 1.8 \\
\midrule
SICK-R-VN & \multicolumn{1}{p{5cm}|}{Một người đàn ông đang ở trong một bãi đậu xe và đang chơi quần vợt với một bức tường lớn.} & \multicolumn{1}{p{5cm}|}{Người trượt tuyết đang nhảy qua tuyết trắng một cách can đảm} & 1.0 \\
\midrule
STSBenchmark-VN & \multicolumn{1}{p{5cm}|}{Người phát ngôn của vận động viên: Các cáo buộc sử dụng doping dường như là không có căn cứ.} & \multicolumn{1}{p{5cm}|}{Tin tức mới nhất về thời tiết khắc nghiệt: 1 người chết ở Texas sau cơn lốc xoáy} & 0.0 \\
\midrule
    \end{tabular}}
    \caption{STS examples. Scores are continuous between 0 and 5 (included).}
    \label{tab:sts_examples}
\end{table*}

\section{Dataset Statistics} \label{appendix:dataset-statistics}
Table \ref{tab:appendix-dataset-detail} provides statistics of all VN-MTEB dataset (after processed and formatted). In our pipeline only the split \texttt{test} is considered to run on the translation verification. 

\label{sec:appdatasets}
\begin{table*}[!t]
    \centering
    \scriptsize
    \resizebox{\textwidth}{!}{\begin{tabular}{l|ccccc}
    \toprule
\bf Name & \bf Type & \bf Train & \bf Dev & \bf Test \\
& & \bf Samples & \bf Samples & \bf Samples \\
\midrule
AmazonCounterfactualVNClassification & Classification & 0 & 0 & 466  \\
AmazonPolarityVNClassification & Classification & 0 & 0 & 344,197  \\
AmazonReviewsVNClassificat,ion & Classification & 0 & 0 & 3,424  \\
Banking77VNClassification & Classification & 0 & 0 & 2,378  \\
EmotionVNClassification & Classification & 0 & 0 & 1,290  \\
ImdbVNClassification & Classification & 0 & 0 & 22,081  \\
MassiveIntentVNClassification & Classification & 0 & 0 & 1784  \\
MassiveScenarioVNClassification & Classification & 0 & 0 & 2974  \\
MTOPDomainVNClassification & Classification & 0 & 0 & 13,291  \\
MTOPIntentVNClassification & Classification & 0 & 0 & 13,291  \\
ToxicConversationsVNClassification & Classification & 0 & 0 & 38,560  \\
TweetSentimentExtractionVNClassification & Classification & 0 & 0 & 2,065  \\
\midrule
RedditClustering-VN & Clustering & 0 & 0 & 293,904  \\
RedditClusteringP2P-VN & Clustering & 0 & 0 & 346,846  \\
StackExchangeClustering-VN & Clustering & 0 & 0 & 251,974  \\
StackExchangeClusteringP2P-VN & Clustering & 0 & 0 & 66,150  \\
TwentyNewsgroupsClustering-VN & Clustering & 0 & 0 & 35,089  \\
\midrule
SprintDuplicateQuestions-VN  & PairClassification & 0 & 0 & 88,173  \\
TwitterSemEval2015-VN  & PairClassification & 0 & 0 & 9,378  \\
TwitterURLCorpus-VN  & PairClassification & 0 & 0 & 30,095  \\
\midrule
AskUbuntuDupQuestions-VN  & Reranking & 0 & 0 & 1,833  \\
SciDocsRR-VN  & Reranking & 0 & 0 & 6,526  \\
StackOverflowDupQuestions-VN  & Reranking & 0 & 0 & 2,808  \\
\midrule
ArguAna-VN & Retrieval & 0 & 0 & 6,969  \\
ClimateFEVER-VN & Retrieval & 0 & 0 & 5,419,992  \\
CQADupstackAndroidRetrieval-VN & Retrieval & 0 & 0 & 24,505  \\
CQADupstackGisRetrieval-VN & Retrieval & 0 & 0 & 38,466  \\
CQADupstackMathematicaRetrieval-VN & Retrieval & 0 & 0 & 17,472  \\
CQADupstackPhysicsRetrieval-VN & Retrieval & 0 & 0 & 39,314  \\
CQADupstackProgrammersRetrieval-VN & Retrieval & 0 & 0 & 33,267  \\
CQADupstackStatsRetrieval-VN & Retrieval & 0 & 0 & 42,693  \\
CQADupstackTexRetrieval-VN & Retrieval & 0 & 0 & 71,313  \\
CQADupstackUnixRetrieval-VN & Retrieval & 0 & 0 & 38,666  \\
CQADupstackWebmastersRetrieval-VN & Retrieval & 0 & 0 & 18,597  \\
CQADupstackWordpressRetrieval-VN & Retrieval & 0 & 0 & 49151  \\
DBPedia-VN & Retrieval & 0 & 0 & 4,540,903  \\
FEVER-VN & Retrieval & 0 & 0 & 5,422,820  \\
FiQA2018-VN & Retrieval & 0 & 0 & 58,659  \\
HotpotQA-VN & Retrieval & 0 & 0 & 5,245,971  \\
MSMARCO-VN & Retrieval & 0 & 0 & 8,846,142  \\
NFCorpus-VN & Retrieval & 0 & 0 & 10,437  \\
NQ-VN & Retrieval & 0 & 0 & 2,683,751  \\
QuoraRetrieval-VN & Retrieval & 0 & 0 & 534,403  \\
SCIDOCS-VN & Retrieval & 0 & 0 & 37,626  \\
SciFact-VN & Retrieval & 0 & 0 & 5,338  \\
Touche2020-VN & Retrieval & 0 & 0 & 383,683  \\
TRECCOVID-VN & Retrieval & 0 & 0 & 228,690  \\
\midrule
BIOSSES-VN & STS & 0 & 0 & 100  \\
SICK-R-VN & STS & 0 & 0 & 9927  \\
STSBenchmark-VN & STS & 0 & 0 & 1379  \\
    \bottomrule
    \end{tabular}}
    \caption{Tasks in VN-MTEB. Dataset already formatted and compatible with MTEB code}
    \label{tab:appendix-dataset-detail}
\end{table*}
\section{Dataset Licenses}
\raggedright
Table \ref{tab:appendix-dataset-licenses} provides publicly available model checkpoints used for VN-MTEB evaluation.
\centering
\begin{table*}[!t]
    \begin{adjustwidth}{0cm}{}
    \scriptsize
    \resizebox{\textwidth}{!}{
    \begin{tabular}{l|l|l|l|c}
    \toprule
    Dataset & Type & Public Link & Translated Link & License \\
    \midrule
    AmazonCounterfactualClassification & Classification & \url{https://huggingface.co/datasets/mteb/amazon_counterfactual} & - & \texttt{cc-by-4.0} \\
    AmazonPolarityClassification & Classification & \url{https://huggingface.co/datasets/mteb/amazon_polarity} & & \texttt{apache-2.0} \\
    AmazonReviewsClassification & Classification & \url{https://huggingface.co/datasets/mteb/amazon_reviews_multi} & - & - \\
    Banking77Classification & Classification & \url{https://huggingface.co/datasets/mteb/banking77} & - & \texttt{mit} \\
    EmotionClassification & Classification & \url{https://huggingface.co/datasets/mteb/emotion} & - & - \\
    ImdbClassification & Classification & \url{https://huggingface.co/datasets/mteb/imdb} & - & - \\
    MassiveIntentClassification & Classification & \url{https://huggingface.co/datasets/mteb/amazon_massive_intent} & - & \texttt{apache-2.0} \\
    MassiveScenarioClassification & Classification & \url{https://huggingface.co/datasets/mteb/amazon_massive_scenario} & - & \texttt{apache-2.0} \\
    MTOPDomainClassification & Classification & \url{https://huggingface.co/datasets/mteb/mtop_domain} & -& - \\
    MTOPIntentClassification & Classification & \url{https://huggingface.co/datasets/mteb/mtop_intent} & - & - \\
    ToxicConversationsClassification & Classification & \url{https://huggingface.co/datasets/mteb/toxic_conversations_50k} & - & \texttt{cc-by-4.0} \\
    TweetSentimentExtractionClassification & Classification & \url{https://huggingface.co/datasets/mteb/tweet_sentiment_extraction} & - & - \\
    \midrule
    RedditClustering & Clustering & \url{https://huggingface.co/datasets/mteb/reddit-clustering} & - & - \\
    RedditClusteringP2P & Clustering & \url{https://huggingface.co/datasets/mteb/reddit-clustering-p2p} & - & - \\
    StackExchangeClustering & Clustering & \url{https://huggingface.co/datasets/mteb/stackexchange-clustering} & - & - \\
    StackExchangeClusteringP2P & Clustering & \url{https://huggingface.co/datasets/mteb/stackexchange-clustering-p2p} & - & - \\
    TwentyNewsgroupsClustering & Clustering & \url{https://huggingface.co/datasets/mteb/twentynewsgroups-clustering} & - & - \\
    \midrule
    SprintDuplicateQuestions & Pair-Classification & \url{https://huggingface.co/datasets/mteb/sprintduplicatequestions-pairclassification} & - & - \\
    TwitterSemEval2015 & Pair-Classification & \url{https://huggingface.co/datasets/mteb/twittersemeval2015-pairclassification} & - & - \\
    TwitterURLCorpus & Pair-Classification & \url{https://huggingface.co/datasets/mteb/twitterurlcorpus-pairclassification} & - & - \\
    \midrule
    AskUbuntuDupQuestions & Reranking & \url{https://huggingface.co/datasets/mteb/askubuntudupquestions-reranking} & - & - \\
    SciDocsRR & Reranking & \url{https://huggingface.co/datasets/mteb/SciDocsRR} & - & \texttt{cc-by-4.0} \\
    StackOverflowDupQuestions & Reranking & \url{https://huggingface.co/datasets/mteb/stackoverflowdupquestions-reranking} & - & - \\
    \midrule
    ArguAna & Retrieval & \url{https://huggingface.co/datasets/mteb/arguana} & - & \texttt{cc-by-4.0} \\
    ClimateFEVER & Retrieval & \url{https://huggingface.co/datasets/mteb/climate-fever} & - & \texttt{cc-by-4.0} \\
    CQADupstackAndroid & Retrieval & \url{https://huggingface.co/datasets/mteb/cqadupstack-android} & - & \texttt{apache-2.0} \\
    CQADupstackGis & Retrieval & \url{https://huggingface.co/datasets/mteb/cqadupstack-gis} & - & \texttt{apache-2.0} \\
    CQADupstackMathematica & Retrieval & \url{https://huggingface.co/datasets/mteb/cqadupstack-mathematica} & - & \texttt{apache-2.0} \\
    CQADupstackPhysics & Retrieval & \url{https://huggingface.co/datasets/mteb/cqadupstack-physics} & - & \texttt{apache-2.0} \\
    CQADupstackProgrammers & Retrieval & \url{https://huggingface.co/datasets/mteb/cqadupstack-programmers} & - &  \texttt{apache-2.0} \\
    CQADupstackStats & Retrieval & \url{https://huggingface.co/datasets/mteb/cqadupstack-stats} & - & \texttt{apache-2.0} \\
    CQADupstackTex & Retrieval & \url{https://huggingface.co/datasets/mteb/cqadupstack-tex} & - & \texttt{apache-2.0} \\
    CQADupstackUnix & Retrieval & \url{https://huggingface.co/datasets/mteb/cqadupstack-unix} & - & \texttt{apache-2.0} \\
    CQADupstackWebmasters & Retrieval & \url{https://huggingface.co/datasets/mteb/cqadupstack-webmasters} & - & \texttt{apache-2.0} \\
    CQADupstackWordpress & Retrieval & \url{https://huggingface.co/datasets/mteb/cqadupstack-wordpress} & - & \texttt{apache-2.0} \\
    DBPedia & Retrieval & \url{https://huggingface.co/datasets/mteb/dbpedia} & - & \texttt{mit} \\
    FEVER & Retrieval & \url{https://huggingface.co/datasets/mteb/fever} & - & \texttt{cc-by-sa-3.0} \\
    FiQA2018 & Retrieval & \url{https://huggingface.co/datasets/mteb/fiqa} & - & \texttt{cc-by-sa-4.0}  \\
    HotpotQA & Retrieval & \url{https://huggingface.co/datasets/mteb/hotpotqa} & - & \texttt{cc-by-sa-4.0} \\
    MSMARCO & Retrieval & \url{https://huggingface.co/datasets/mteb/msmarco} & - & \texttt{cc-by-sa-4.0}  \\
    NFCorpus & Retrieval & \url{https://huggingface.co/datasets/mteb/nfcorpus} & - & \texttt{cc-by-sa-4.0}  \\
    NQ & Retrieval & \url{https://huggingface.co/datasets/mteb/nq} & - & \texttt{cc-by-nc-sa-3.0} \\
    Quora & Retrieval & \url{https://huggingface.co/datasets/mteb/quora} & - & \texttt{cc-by-sa-4.0}  \\
    SCIDOCS & Retrieval & \url{https://huggingface.co/datasets/mteb/scidocs} & - & \texttt{cc-by-sa-4.0} \\
    SciFact & Retrieval & \url{https://huggingface.co/datasets/mteb/scifact} & - & \texttt{cc-by-sa-4.0}  \\
    Touche2020 & Retrieval & \url{https://huggingface.co/datasets/mteb/touche2020} & - & \texttt{cc-by-sa-4.0}  \\
    TRECCOVID & Retrieval & \url{https://huggingface.co/datasets/mteb/trec-covid} & - & \texttt{cc-by-sa-4.0} \\
    \midrule
    BIOSSES & STS & \url{https://huggingface.co/datasets/mteb/biosses-sts} & - & - \\
    SICK-R & STS & \url{https://huggingface.co/datasets/mteb/sickr-sts} & - & \texttt{cc-by-nc-sa-3.0} \\
    STSBenchmark & STS & \url{https://huggingface.co/datasets/mteb/stsbenchmark-sts} & - & - \\
    \midrule
    \bottomrule
    \end{tabular}
    }
    \end{adjustwidth}
    \caption{Dataset licenses for MTEB and VN-MTEB}
    \label{tab:appendix-dataset-licenses}
\end{table*}

\section{GPU usage for translation} \label{appendix:gpu-usage}
\begin{table*}[!t]
    \centering
    \scriptsize
    \resizebox{\textwidth}{!}{\begin{tabular}{l|ccccc}
    \toprule
\bf Name & \bf Type & \bf Total Number of tokens & \bf Time Estimated (s) & \bf GPU Electricity Consumption (kWh) \\
\midrule
\midrule
AmazonCounterfactualVNClassification & Classification & 910,364 & 239.57 & 0.186  \\
AmazonPolarityVNClassification & Classification & 536,435,795 & 141167.31 & 109.797  \\
AmazonReviewsVNClassification & Classification & 82,306,198 & 21659.53 & 16.846  \\
Banking77VNClassification & Classification & 241,685 & 63.60 & 0.049  \\
EmotionVNClassification & Classification & 595,593 & 156.74 & 0.122  \\
ImdbVNClassification & Classification & 18,074,863 & 4756.54 & 3.700  \\
MassiveIntentVNClassification & Classification & 13,809,421 & 3634.06 & 2.826  \\
MassiveScenarioVNClassification & Classification & 13,802,417 & 3632.22 & 2.825  \\
MTOPDomainVNClassification & Classification & 1,439,620 & 378.85 & 0.295  \\
MTOPIntentVNClassification & Classification & 1,439,620 & 378.85 & 0.295  \\
ToxicConversationsVNClassification & Classification & 9,332,763 & 2455.99 & 1.910  \\
TweetSentimentExtractionVNClassification & Classification & 1,011,699 & 266.24 & 0.207  \\
\midrule
RedditClustering-VN & Clustering & 12,694,431 & 3340.64 & 2.598  \\
RedditClusteringP2P-VN & Clustering & 108,712,751 & 28608.62 & 22.251  \\
StackExchangeClustering-VN & Clustering & 17,157,163 & 4515.04 & 3.512  \\
StackExchangeClusteringP2P-VN & Clustering & 25,618,672 & 6741.76 & 5.244  \\
TwentyNewsgroupsClustering-VN & Clustering & 1,655,500 & 435.66 & 0.339  \\
\midrule
SprintDuplicateQuestions-VN & PairClassification & 4,711,640 & 1239.91 & 0.964  \\
TwitterSemEval2015-VN & PairClassification & 665,973 & 175.26 & 0.136  \\
TwitterURLCorpus-VN & PairClassification & 3,004,908 & 790.77 & 0.615  \\
\midrule
AskUbuntuDupQuestions-VN & Reranking & 136,142 & 35.83 & 0.028  \\
SciDocsRR-VN & Reranking & 7,620,209 & 2005.32 & 1.560  \\
StackOverflowDupQuestions-VN & Reranking & 12,324,554 & 3243.30 & 2.523  \\
\midrule
ArguAna-VN & Retrieval & 2,842,260 & 747.96 & 0.582  \\
ClimateFEVER-VN & Retrieval & 681,973,189 & 179466.63 & 139.585  \\
CQADupstackAndroidRetrieval-VN & Retrieval & 3,902,043 & 1026.85 & 0.799  \\
CQADupstackGisRetrieval-VN & Retrieval & 10,313,933 & 2714.19 & 2.111  \\
CQADupstackMathematicaRetrieval-VN & Retrieval & 6,109,244 & 1607.70 & 1.250  \\
CQADupstackPhysicsRetrieval-VN & Retrieval & 6,224,273 & 1637.97 & 1.274  \\
CQADupstackProgrammersRetrieval-VN & Retrieval & 8,800,245 & 2315.85 & 1.801  \\
CQADupstackStatsRetrieval-VN & Retrieval & 13,178,147 & 3467.93 & 2.697  \\
CQADupstackTexRetrieval-VN & Retrieval & 25,201,127 & 6631.88 & 5.158  \\
CQADupstackUnixRetrieval-VN & Retrieval & 13,401,968 & 3526.83 & 2.743  \\
CQADupstackWebmastersRetrieval-VN & Retrieval & 3,483,317 & 916.66 & 0.713  \\
CQADupstackWordpressRetrieval-VN & Retrieval & 14,241,887 & 3747.86 & 2.915  \\
DBPedia-VN & Retrieval & 414,726,629 & 109138.59 & 84.886  \\
FEVER-VN & Retrieval & 683,783,334 & 179942.98 & 139.956  \\
FiQA2018-VN & Retrieval & 12,536,252 & 3299.01 & 2.566  \\
HotpotQA-VN & Retrieval & 442,305,098 & 116396.08 & 90.530  \\
MSMARCO-VN & Retrieval & 778,538,066 & 204878.44 & 159.350  \\
NFCorpus-VN & Retrieval & 1,642,900 & 432.34 & 0.336  \\
NQ-VN & Retrieval & 370,480,772 & 97494.94 & 75.829  \\
QuoraRetrieval-VN & Retrieval & 19,285,282 & 5075.07 & 3.947  \\
SCIDOCS-VN & Retrieval & 7,936,076 & 2088.44 & 1.624  \\
SciFact-VN & Retrieval & 2,200,704 & 579.13 & 0.450  \\
Touche2020-VN & Retrieval & 170,315,421 & 44819.85 & 34.860  \\
TRECCOVID-VN & Retrieval & 52,994,734 & 13945.98 & 10.847  \\
\midrule
BIOSSES-VN & STS & 9,357 & 2.46 & 0.002  \\
SICK-R-VN & STS & 269,368 & 70.89 & 0.055  \\
STSBenchmark-VN & STS & 332,610 & 87.53 & 0.068  \\
\midrule
Total & Total & 4,620,730,217 & 1215981.64 & 946.066 \\
    \bottomrule
    
    \end{tabular}}
    \caption{GPU Usage to Translate datasets in VN-MTEB}
    \label{tab:gpu-usage-simplified}
\end{table*}
\raggedright
In our experiment, we utilized 4 H100 GPUs, each GPU electricity consumption is about 700W. As shown in Table \ref{tab:gpu-usage-simplified}, we measured an output token rate of 3,800 tokens per second. Since the entire process requires counting both input and output tokens, we multiply this rate by $2$ to accurately estimate the time and energy consumption for each dataset as well as the overall workload.
To summary, the estimated time to translate all VN-MTEB dataset is 
\[
\begin{aligned}
\text{Total time} \times 2 &= 1,215,981.64 \text{ seconds} \times 2 \\
&= 2,431,963.28 \text{ seconds} \\
&\approx 675.54 \text{ hours} \\
&\approx 28.14 \text{ days}
\end{aligned}
\]
\section{Model performance with size} 
\begin{figure*}
  \includegraphics[width=\textwidth, scale=0.55]{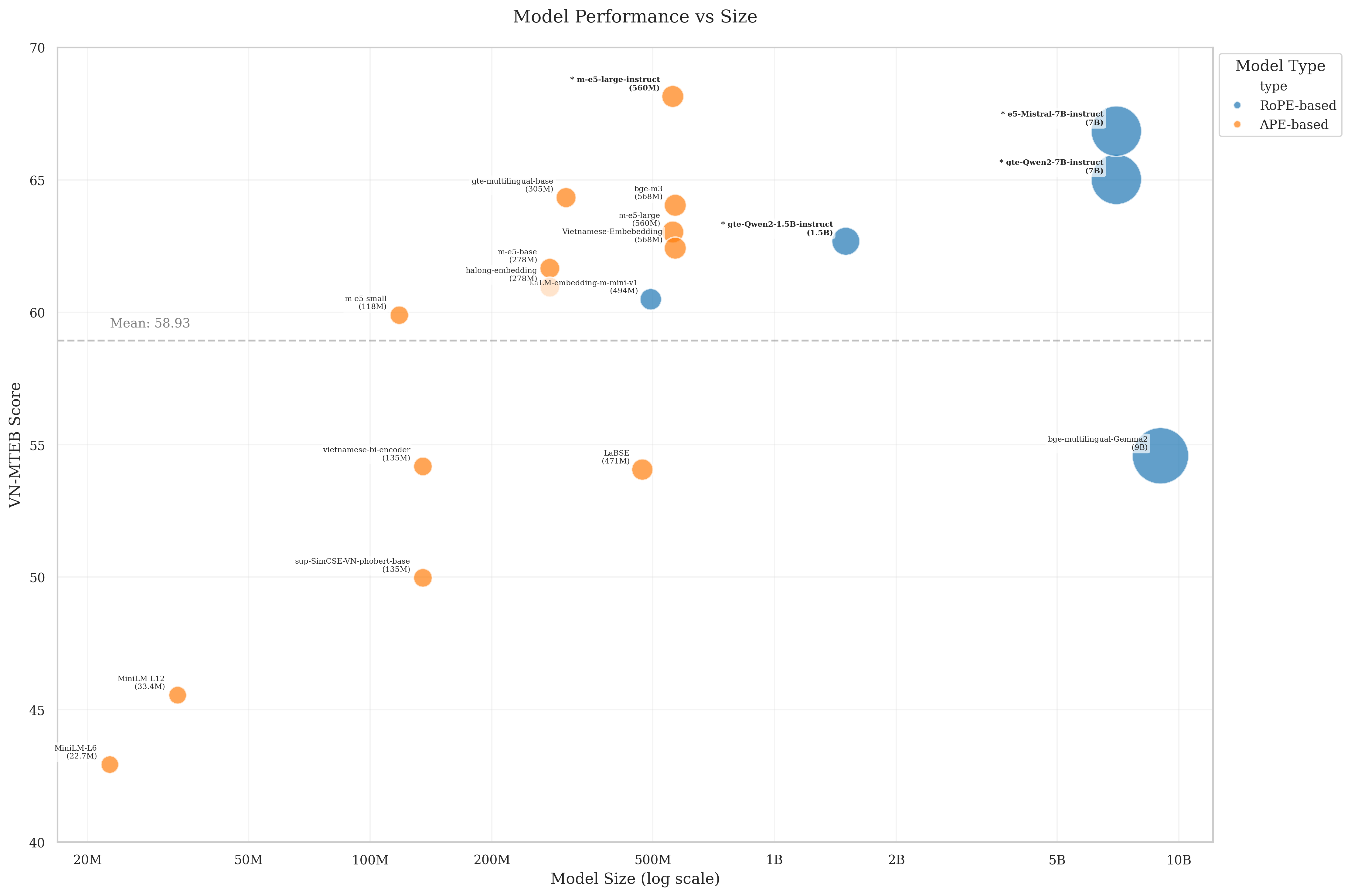}
  \caption{Model performance and size.}
  \label{fig:model-performance-and-size}
\end{figure*}

Figure \ref{fig:model-performance-and-size} represent an overview of model performance along with size and model type.
\section{Model}
Table \ref{tab:appendix-ckpts} provides publicly available model checkpoints used for MTEB evaluation.
\begin{table*}[!t]
    \centering
    \tiny
    \resizebox{\textwidth}{!}{\begin{tabular}{l|l}
    \toprule
\multicolumn{1}{l|}{\textbf{Model}} & \multicolumn{1}{c}{\textbf{Public Checkpoint}} \\
\midrule
gte-Qwen2-7B-instruct & \url{https://huggingface.co/Alibaba-NLP/gte-Qwen2-7B-instruct} \\
e5-Mistral-7B-instruct & \url{https://huggingface.co/intfloat/e5-mistral-7b-instruct} \\
bge-multilingual-Gemma2 & \url{https://huggingface.co/BAAI/bge-multilingual-gemma2} \\
gte-Qwen2-1.5B-instruct & \url{https://huggingface.co/Alibaba-NLP/gte-Qwen2-1.5B-instruct} \\
m-e5-large-instruct & \url{https://huggingface.co/intfloat/multilingual-e5-large-instruct} \\
m-e5-large & \url{https://huggingface.co/intfloat/multilingual-e5-large} \\
bge-me & \url{https://huggingface.co/BAAI/bge-m3} \\
Vietnamese-Embedding & \url{https://huggingface.co/AITeamVN/Vietnamese_Embedding} \\
KaLM-embedding-m-mini-v1 & \url{https://huggingface.co/HIT-TMG/KaLM-embedding-multilingual-mini-v1} \\
LaBSE & \url{https://huggingface.co/sentence-transformers/LaBSE} \\
gte-multilingual-base & \url{https://huggingface.co/Alibaba-NLP/gte-multilingual-base} \\
m-e5-base & \url{https://huggingface.co/intfloat/multilingual-e5-base} \\
halong-embedding  & \url{https://huggingface.co/hiieu/halong_embedding} \\
m-e5-small  & \url{https://huggingface.co/intfloat/multilingual-e5-small} \\
vietnamese-bi-encoder  & \url{https://huggingface.co/bkai-foundation-models/vietnamese-bi-encoder} \\
sup-SimCSE-VN-phobert-base   & \url{https://huggingface.co/VoVanPhuc/sup-SimCSE-VietNamese-phobert-base} \\
MiniLM-L12   & \url{https://huggingface.co/sentence-transformers/paraphrase-multilingual-MiniLM-L12-v2} \\
MiniLM-L6  & \url{https://huggingface.co/sentence-transformers/paraphrase-multilingual-MiniLM-L6-v2} \\
    \bottomrule
    \end{tabular}}
    \caption{Publicly available model links used for evaluation}
    \label{tab:appendix-ckpts}
\end{table*}

\section{Detail Model Result}\label{appendix:detail-model-result}
Table \ref{tab:detail-ape-results} and table \ref{tab:detail-rope-results} represent detail model result. We split into 2 tables, each for RoPE-based and other one is for APE-based.

\pagenumbering{gobble}
\begin{landscape}
    \centering
    \begin{table*}[!t]
        \begin{adjustwidth}{-4cm}{}
        \tiny
        \resizebox{\textwidth}{!}{
        \begin{tabular}{l|ccccc}
        \toprule
        Dataset & gte-Qwen2-7B-instruct & e5-Mistral-7B-instruct & bge-multilingual-Gemma2 & gte-Qwen2-1.5B-instruct & KaLM-mini \\
        \midrule
        AmazonCounterfactualVNClassification & 66.7 & 68.8 & 68.78 & 64.48 & 62.36 \\
        AmazonPolarityVNClassification & 90.89 & 93.8 & 84.14 & 82.0 & 75.84 \\
        AmazonReviewsVNClassification & 43.23 & 49.94 & 42.03 & 38.71 & 40.05 \\
        Banking77VNClassification & 83.04 & 83.86 & 83.88 & 81.88 & 71.63 \\
        EmotionVNClassification & 46.19 & 44.8 & 50.23 & 45.16 & 43.13 \\
        ImdbVNClassification & 86.63 & 88.09 & 81.51 & 70.43 & 73.12 \\
        MassiveIntentVNClassification & 74.34 & 75.8 & 72.59 & 72.37 & 63.55 \\
        MassiveScenarioVNClassification & 78.28 & 78.74 & 76.48 & 75.88 & 67.37 \\
        MTOPDomainVNClassification & 89.62 & 88.43 & 91.66 & 86.99 & 81.04 \\
        MTOPIntentVNClassification & 70.43 & 68.7 & 75.72 & 66.48 & 53.63 \\
        ToxicConversationsVNClassification & 61.22 & 62.35 & 73.19 & 60.74 & 62.49 \\
        TweetSentimentExtractionVNClassification & 58.52 & 63.27 & 61.13 & 60.56 & 59.85 \\
        \midrule
        RedditClustering-VN & 49.7 & 45.78 & 29.91 & 46.76 & 45.37 \\
        RedditClusteringP2P-VN & 64.06 & 59.34 & 56.5 & 56.65 & 60.68 \\
        StackExchangeClustering-VN & 65.05 & 62.72 & 48.83 & 58.9 & 55.67 \\
        StackExchangeClusteringP2P-VN & 40.67 & 43.8 & 32.99 & 33.42 & 33.37 \\
        TwentyNewsgroupsClustering-VN & 46.27 & 46.9 & 32.42 & 42.46 & 39.16 \\
        \midrule
        SprintDuplicateQuestions-VN & 75.07 & 91.78 & 66.68 & 85.03 & 90.6 \\
        TwitterSemEval2015-VN & 58.68 & 73.32 & 53.76 & 52.44 & 63.65 \\
        TwitterURLCorpus-VN & 82.52 & 86.92 & 80.49 & 80.64 & 85.58 \\
        \midrule
        AskUbuntuDupQuestions-VN & 77.03 & 78.17 & 68.05 & 73.01 & 70.93 \\
        SciDocsRR-VN & 93.62 & 93.32 & 83.93 & 92.18 & 90.12 \\
        StackOverflowDupQuestions-VN & 52.2 & 53.96 & 40.63 & 48.91 & 45.48 \\
        \midrule
        ArguAna-VN & 52.77 & 50.36 & 50.61 & 51.99 & 52.66 \\
        ClimateFEVER-VN & 21.49 & 24.77 & 16.52 & 23.47 & 7.81 \\
        CQADupstackAndroid-VN & 48.36 & 46.82 & 34.54 & 42.33 & 43.3 \\
        CQADupstackGis-VN & 36.06 & 35.18 & 15.15 & 28.13 & 29.8 \\
        CQADupstackMathematica-VN & 29.41 & 25.26 & 12.22 & 24.46 & 20.73 \\
        CQADupstackPhysics-VN & 48.15 & 38.17 & 24.0 & 37.18 & 36.64 \\
        CQADupstackProgrammers-VN & 38.86 & 40.42 & 19.15 & 35.66 & 33.66 \\
        CQADupstackStats-VN & 34.59 & 29.55 & 10.96 & 26.77 & 26.69 \\
        CQADupstackTex-VN & 26.74 & 28.1 & 8.66 & 23.75 & 23.29 \\
        CQADupstackUnix-VN & 39.26 & 39.94 & 20.01 & 33.88 & 32.97 \\
        CQADupstackWebmasters-VN & 38.71 & 38.59 & 20.35 & 32.3 & 32.5 \\
        CQADupstackWordpress-VN & 31.14 & 31.62 & 11.45 & 25.34 & 23.55 \\
        DBPedia-VN & 41.89 & 42.78 & 6.96 & 39.51 & 28.61 \\
        FEVER-VN & 82.81 & 84.82 & 45.23 & 83.53 & 60.61 \\
        FiQA2018-VN & 46.92 & 30.39 & 11.76 & 34.27 & 29.45 \\
        HotpotQA-VN & 67.99 & 64.54 & 29.72 & 61.86 & 60.81 \\
        MSMARCO-VN & 68.99 & 35.24 & 10.3 & 66.49 & 28.31 \\
        NFCorpus-VN & 38.27 & 31.98 & 10.25 & 33.21 & 29.76 \\
        NQ-VN & 59.91 & 57.8 & 9.71 & 54.89 & 34.42 \\
        Quora-VN & 52.23 & 42.87 & 21.3 & 52.11 & 52.14 \\
        SCIDOCS-VN & 20.95 & 15.23 & 8.12 & 18.04 & 13.83 \\
        SciFact-VN & 73.8 & 63.77 & 45.29 & 69.67 & 58.74 \\
        Touche2020-VN & 28.64 & 25.92 & 11.05 & 30.99 & 22.17 \\
        TRECCOVID-VN & 77.3 & 77.42 & 39.2 & 78.46 & 59.33 \\
        \midrule
        BIOSSES-VN & 82.09 & 83.72 & 66.85 & 80.8 & 83.52 \\
        SICK-R-VN & 76.32 & 77.91 & 66.5 & 78.07 & 74.49 \\
        STSBenchmark-VN & 77.79 & 81.98 & 64.97 & 81.03 & 77.6 \\
        \midrule
        \end{tabular}
        }
        \end{adjustwidth}
        \begin{adjustwidth}{-8.5cm}{}
        \caption{All Vietnamese results on RoPE based model. The main score for each task is reported as described in Original MTEB 
        Paper \cite{muennighoff-etal-2023-mteb}.}
        \label{tab:detail-rope-results}
        \end{adjustwidth}
    \end{table*}
    
\end{landscape}
\begin{landscape}
    \begin{table*}[h]
        \begin{adjustwidth}{-9cm}{}
        \centering
        \tiny
        \resizebox{1.6\textwidth}{!}{
        \begin{tabular}{l|ccccccccccccc}
        \toprule
        Dataset & m-e5-large-instruct & m-e5-large & bge-m3 & Vietnamese-Emb & LaBSE & gte-multilingual-base & m-e5-base & halong & m-e5-small & vietnamese-bi & sup-SimCSE-VN & MiniLM-L12 & MiniLM-L6 \\
        \midrule
        AmazonCounterfactualVNClassification & 67.7 & 70.39 & 69.4 & 71.44 & 71.61 & 66.24 & 66.09 & 65.6 & 63.07 & 61.37 & 67.96 & 64.7 & 64.59 \\
        AmazonPolarityVNClassification & 95.05 & 76.42 & 87.54 & 88.78 & 70.39 & 80.06 & 75.91 & 69.99 & 74.86 & 66.52 & 79.05 & 55.4 & 56.19 \\
        AmazonReviewsVNClassification & 49.8 & 39.68 & 44.33 & 44.48 & 36.37 & 42.36 & 40.31 & 36.36 & 38.55 & 32.79 & 37.69 & 27.22 & 26.99 \\
        Banking77VNClassification & 83.84 & 73.74 & 78.1 & 79.21 & 67.11 & 74.71 & 70.96 & 75.02 & 67.14 & 75.51 & 69.05 & 50.8 & 48.94 \\
        EmotionVNClassification & 49.64 & 46.81 & 49.93 & 48.95 & 40.91 & 44.31 & 45.24 & 46.85 & 40.5 & 34.57 & 36.69 & 20.14 & 20.96 \\
        ImdbVNClassification & 91.92 & 72.68 & 82.71 & 83.06 & 62.59 & 75.31 & 68.51 & 65.2 & 66.77 & 59.01 & 69.93 & 51.53 & 54.24 \\
        MassiveIntentVNClassification & 74.38 & 65.73 & 68.18 & 67.74 & 60.59 & 64.96 & 63.02 & 65.4 & 60.06 & 62.29 & 57.94 & 42.39 & 41.47 \\
        MassiveScenarioVNClassification & 77.62 & 68.32 & 72.75 & 72.85 & 64.1 & 69.37 & 67.24 & 70.88 & 64.38 & 65.36 & 60.76 & 47.84 & 45.9 \\
        MTOPDomainVNClassification & 87.74 & 84.75 & 86.56 & 85.54 & 79.72 & 82.82 & 83.98 & 84.29 & 79.35 & 79.35 & 70.58 & 58.9 & 56.41 \\
        MTOPIntentVNClassification & 71.92 & 57.33 & 57.01 & 58.01 & 53.23 & 50.94 & 52.01 & 53.99 & 45.5 & 55.84 & 48.21 & 30.43 & 29.93 \\
        ToxicConversationsVNClassification & 67.03 & 64.22 & 69.08 & 66.27 & 65.05 & 68.67 & 65.67 & 66.59 & 63.34 & 62.94 & 61.46 & 55.56 & 54.75 \\
        TweetSentimentExtractionVNClassification & 64.02 & 60.25 & 63.49 & 62.34 & 59.53 & 60.11 & 60.56 & 59.77 & 59.69 & 51.45 & 56.92 & 41.89 & 41.89 \\
        \midrule
        RedditClustering-VN & 49.06 & 42.12 & 43.25 & 43.89 & 28.29 & 49.91 & 42.6 & 38.31 & 37.74 & 28.6 & 29.08 & 17.97 & 13.23 \\
        RedditClusteringP2P-VN & 61.47 & 60.64 & 57.38 & 56.16 & 50.09 & 59.75 & 58.34 & 55.67 & 56.39 & 43.82 & 43.66 & 31.45 & 27.61 \\
        StackExchangeClustering-VN & 63.63 & 56.39 & 58.42 & 57.24 & 38.58 & 60.8 & 57.15 & 55.26 & 54.88 & 44.31 & 38.63 & 21.91 & 16.62 \\
        StackExchangeClusteringP2P-VN & 41.26 & 32.07 & 32.63 & 31.49 & 27.87 & 35.23 & 32.22 & 31.94 & 32.5 & 27.8 & 27.66 & 29.49 & 24.89 \\
        TwentyNewsgroupsClustering-VN & 49.4 & 37.69 & 37.83 & 39.29 & 28.11 & 45.55 & 38.2 & 35.92 & 34.31 & 26.12 & 26.2 & 20.98 & 19.66 \\
        \midrule
        SprintDuplicateQuestions-VN & 90.27 & 93.58 & 96.54 & 95.28 & 82.55 & 97.08 & 93.16 & 95.23 & 91.42 & 89.14 & 76.27 & 80.74 & 69.59 \\
        TwitterSemEval2015-VN & 76.1 & 71.78 & 70.99 & 68.24 & 65.26 & 70.21 & 68.76 & 64.02 & 67.47 & 61.05 & 57.67 & 50.2 & 52.33 \\
        TwitterURLCorpus-VN & 87.03 & 85.75 & 85.77 & 84.99 & 84.91 & 85.99 & 85.62 & 84.36 & 84.66 & 82.02 & 79.99 & 77.45 & 76.46 \\
        \midrule
        AskUbuntuDupQuestions-VN & 75.09 & 71.39 & 72.73 & 72.97 & 67.88 & 73.23 & 69.99 & 70.51 & 68.5 & 68.0 & 62.43 & 66.57 & 65.84 \\
        SciDocsRR-VN & 93.34 & 91.1 & 90.01 & 88.77 & 84.72 & 91.83 & 89.52 & 88.91 & 88.2 & 83.01 & 78.88 & 75.74 & 74.62 \\
        StackOverflowDupQuestions-VN & 51.41 & 48.72 & 51.09 & 50.94 & 44.33 & 50.29 & 47.69 & 50.08 & 46.36 & 43.83 & 35.28 & 39.01 & 37.91 \\
        \midrule
        ArguAna-VN & 48.15 & 47.88 & 50.68 & 51.07 & 36.46 & 52.75 & 45.49 & 52.48 & 42.97 & 38.08 & 26.35 & 9.89 & 9.72 \\
        ClimateFEVER-VN & 25.01 & 15.43 & 21.27 & 13.25 & 2.41 & 21.05 & 12.62 & 14.48 & 15.13 & 11.14 & 7.0 & 1.63 & 0.4 \\
        CQADupstackAndroid-VN & 43.13 & 42.28 & 44.04 & 41.93 & 28.47 & 39.66 & 42.35 & 42.12 & 41.69 & 26.73 & 16.56 & 20.84 & 17.25 \\
        CQADupstackGis-VN & 30.73 & 31.28 & 33.13 & 31.91 & 17.24 & 29.12 & 28.61 & 30.76 & 29.12 & 17.8 & 8.28 & 13.8 & 9.19 \\
        CQADupstackMathematica-VN & 22.31 & 24.06 & 23.64 & 21.44 & 12.85 & 20.8 & 21.33 & 21.85 & 19.33 & 13.19 & 4.54 & 9.72 & 6.62 \\
        CQADupstackPhysics-VN & 35.7 & 36.53 & 37.99 & 35.52 & 21.19 & 39.08 & 35.15 & 36.89 & 36.96 & 26.19 & 14.16 & 14.54 & 10.19 \\
        CQADupstackProgrammers-VN & 36.74 & 34.53 & 34.12 & 32.71 & 18.51 & 34.19 & 31.9 & 32.85 & 31.42 & 20.42 & 10.74 & 14.72 & 7.77 \\
        CQADupstackStats-VN & 26.19 & 27.81 & 30.12 & 26.86 & 15.08 & 27.79 & 25.81 & 28.57 & 26.51 & 18.64 & 7.3 & 14.8 & 7.48 \\
        CQADupstackTex-VN & 23.4 & 22.68 & 26.11 & 24.78 & 12.73 & 21.37 & 20.78 & 23.97 & 22.08 & 10.99 & 5.59 & 10.53 & 6.07 \\
        CQADupstackUnix-VN & 32.98 & 33.62 & 35.67 & 34.52 & 22.5 & 30.61 & 32.94 & 32.65 & 31.12 & 19.48 & 8.82 & 16.44 & 11.78 \\
        CQADupstackWebmasters-VN & 33.85 & 33.07 & 34.47 & 31.67 & 20.78 & 28.51 & 31.04 & 32.6 & 30.58 & 21.39 & 11.41 & 16.52 & 9.56 \\
        CQADupstackWordpress-VN & 25.3 & 25.56 & 28.18 & 24.74 & 14.05 & 23.38 & 23.87 & 24.78 & 23.39 & 16.21 & 6.45 & 13.23 & 8.54 \\
        DBPedia-VN & 39.9 & 31.58 & 36.7 & 34.2 & 15.92 & 37.46 & 30.77 & 23.8 & 28.54 & 20.22 & 11.16 & 14.81 & 10.9 \\
        FEVER-VN & 83.34 & 58.3 & 70.14 & 48.81 & 12.58 & 86.24 & 49.6 & 52.87 & 54.25 & 53.11 & 11.89 & 29.4 & 12.35 \\
        FiQA2018-VN & 36.46 & 31.51 & 34.38 & 29.94 & 7.38 & 32.88 & 25.14 & 26.23 & 22.71 & 17.29 & 6.62 & 4.31 & 1.44 \\
        HotpotQA-VN & 63.99 & 65.11 & 63.76 & 70.07 & 17.0 & 58.6 & 60.79 & 53.36 & 54.99 & 34.48 & 13.65 & 17.16 & 13.31 \\
        MSMARCO-VN & 37.86 & 39.08 & 36.22 & 30.5 & 10.12 & 35.16 & 36.19 & 29.75 & 33.11 & 30.12 & 4.99 & 9.41 & 8.16 \\
        NFCorpus-VN & 33.4 & 31.5 & 30.96 & 25.39 & 20.5 & 31.48 & 26.75 & 27.03 & 27.28 & 23.38 & 15.82 & 17.32 & 14.05 \\
        NQ-VN & 56.86 & 52.32 & 54.98 & 42.61 & 11.7 & 50.65 & 45.1 & 36.15 & 38.54 & 30.63 & 7.08 & 10.79 & 7.44 \\
        Quora-VN & 57.9 & 66.49 & 64.57 & 61.0 & 38.17 & 56.68 & 63.29 & 58.79 & 60.47 & 37.55 & 32.33 & 26.55 & 20.43 \\
        SCIDOCS-VN & 16.81 & 13.74 & 15.01 & 13.03 & 8.36 & 14.49 & 12.9 & 13.35 & 11.71 & 9.18 & 4.93 & 5.39 & 3.61 \\
        SciFact-VN & 65.52 & 68.5 & 62.31 & 55.12 & 41.49 & 65.62 & 67.61 & 60.87 & 65.78 & 39.29 & 19.67 & 26.75 & 20.93 \\
        Touche2020-VN & 25.03 & 16.01 & 21.53 & 11.98 & 3.93 & 22.69 & 13.13 & 15.88 & 17.74 & 18.8 & 12.15 & 2.68 & 2.66 \\
        TRECCOVID-VN & 80.56 & 54.71 & 66.22 & 27.32 & 16.96 & 60.82 & 44.86 & 54.73 & 53.57 & 54.57 & 21.23 & 18.11 & 11.75 \\
        \midrule
        BIOSSES-VN & 84.26 & 81.69 & 77.5 & 78.14 & 76.77 & 84.45 & 81.82 & 80.2 & 79.08 & 66.13 & 55.13 & 64.14 & 56.09 \\
        SICK-R-VN & 80.17 & 78.22 & 77.88 & 77.11 & 68.77 & 77.5 & 76.77 & 74.0 & 75.49 & 69.65 & 74.46 & 61.92 & 62.05 \\
        STSBenchmark & 84.38 & 82.03 & 81.15 & 77.19 & 70.59 & 82.58 & 79.77 & 77.96 & 78.09 & 69.97 & 76.24 & 60.94 & 56.62 \\
        \midrule
        \bottomrule
        \end{tabular}
        }
        \end{adjustwidth}
        \begin{adjustwidth}{-8cm}{}
        \caption{All Vietnamese results on APE based model. The main score for each task is reported as described in Original MTEB paper \cite{muennighoff-etal-2023-mteb}.}
        \label{tab:detail-ape-results}
        \end{adjustwidth}
    \end{table*}
    \end{landscape}
\pagenumbering{arabic}

\end{document}